\documentclass[journal]{IEEEtran}
\usepackage{amssymb}
\usepackage{mathrsfs}
\usepackage{stmaryrd}
\usepackage{amsfonts}
\usepackage{latexsym}
\usepackage{epsfig}
\usepackage{listings}
\usepackage{booktabs}
\usepackage{subfigure}
\usepackage{amsmath,bm}
\usepackage{picins}

\usepackage{algorithm}
\usepackage{algorithmic}
\usepackage{multirow}
\usepackage{amsfonts}

\ifCLASSINFOpdf
\else
\fi
\begin{document}
%
\title{Semi-supervised Sparse Representation with Graph Regularization for Image Classification}
%
%
%

\author{Hongfeng~Li
\thanks{This work was supported in part by the National Natural Science Foundation of China under Grant 11701018.}
\thanks{H.~F.~Li is with Center for Data Science in Health and Medicine, Peking University, Beijing, China (e-mail: lihongfeng2010@gmail.com).}
}
\maketitle

\begin{abstract}
Image classification is a challenging problem for computer in reality. Large numbers of methods can achieve satisfying performances with sufficient labeled images. However, labeled images are still highly limited for certain image classification tasks. Instead, lots of unlabeled images are available and easy to be obtained. Therefore, making full use of the available unlabeled data can be a potential way to further improve the performance of current image classification methods. In this paper, we propose a discriminative semi-supervised sparse representation algorithm for image classification. In the algorithm, the classification process is combined with the sparse coding to learn a data-driven linear classifier. To obtain discriminative predictions, the predicted labels are regularized with three graphs, i.e., the global manifold structure graph, the within-class graph and the between-classes graph. The constructed graphs are able to extract structure information included in both the labeled and unlabeled data. Moreover, the proposed method is extended to a kernel version for dealing with data that cannot be linearly classified. Accordingly, efficient algorithms are developed to solve the corresponding optimization problems. Experimental results on several challenging databases demonstrate that the proposed algorithm achieves excellent performances compared with related popular methods.
\end{abstract}


%
\IEEEpeerreviewmaketitle

\section{Introduction}
%
%
%
%
\IEEEPARstart{W}{ith} the rapid development of technology, overwhelming tons of images are available to us, which makes it necessary for us to develop novel image classification algorithms to choose the desired ones efficiently. However, images are commonly of high dimensionality and vary greatly with various scales, lighting conditions, viewpoints and context variations, making it challenging for image classification tasks. Particularly, feature extraction is an essential process for image classification tasks. Therefore, how to extract effective features from images has become a crucial problem for image classification.

As a powerful tool for extracting essential features and obtaining high-level semantics from images, sparse coding (SC) methods \cite{hoyer2002non,lee2006efficient} try to represent data with only a few non-zero sparse coefficients. To obtain the sparse representation of the data, we commonly minimize the distance between the data and the linear combination of a dictionary, and enforce sparsity constraint to the coefficients. SC assumes that a few atoms in the dictionary are enough for representing the data, i.e., most of the coefficients are zeros while only a few are non-zeros. The sparse representation is easy to be interpreted and robust to noise, hence, has been a popular and effective method for feature extraction in image classification problems \cite{mairal2009online,zheng2011graph,yang2009linear,li2014sparse,li2018neural}.

In spite of the extensive applications of SC, such as object recognition \cite{yang2009linear,li2013hierarchical}, face recognition \cite{li2014sparse,wright2009robust,aharon2006img,jiang2013label}, human action recognition \cite{guha2012learning} and digit recognition \cite{shrivastava2015multiple}, the SC algorithms are mostly utilized in an unsupervised way. In these cases, the available label information is ignored when the SC is employed to extract features from images. Specifically, for image classification problems, only the representations of the images are input to a classifier for classification and the available labels of the images are not involved. Consequently, the discriminative ability of SC could be further improved if the available label information can be utilized properly.

Supervised SC algorithms have been proposed to make full use of the label information during the sparse representation learning process. For instance, the authors in \cite{mairal2009supervised} and \cite{zhang2010discriminative} unified the sparse coding and classifier training processes in one objective function to learn the sparse representations of data and data-driven classifier simultaneously. LC-KSVD \cite{jiang2013label} learned a discriminative dictionary for sparse coding by associating the label information of the training data to each atom in the dictionary and enforcing a label consistency constraint. Despite the success of the supervised methods, they always require that the whole training data be labeled or the number of labeled data be relatively large. Nevertheless, massive labeled data is impractical in real-world applications as labeling data is typically difficult and expensive. The supervised methods cannot work well when the labeled data is insufficient. By contrast, it is much easier to obtain unlabeled data and partially labeled databases are more ubiquitous in practice. Consequently, utilizing the information included in the unlabeled data can be beneficial for learning methods.

Semi-supervised learning (SSL) methods that learn features from both the labels of the data and the overall distribution of the labeled and unlabeled data provide an alternative way to alleviate the above practical problems \cite{chapelle2006semi,cai2007semi,yang2012semi}. In the case that only a small amount of labeled data is available for training, supervised learning methods may fail to learn effective features. By contrast, SSL methods can exploit the beneficial information contained in the unlabeled data, thus achieve better performances. Therefore, developing methods to learn information from both labeled and unlabeled data is more desirable. Many SSL methods have been proposed to learn classifier or the labels of the unlabeled data by utilizing both the labeled and unlabeled data recently \cite{wang2014semi,yu2006active,he2011nonnegative,zheng2013image}. Obviously, successful methods should be able to utilize the available labels of the data and feature vectors of the labeled and unlabeled data simultaneously to learn discriminative features. Moreover, the desirable features should be as smooth as the original data in the subspace while possess discriminative properties for classification. However, few sparse coding based SSL methods can meet the requirement of neighborhood smoothness and being discriminative among different labeled data at the same time.

To resolve the above issues, in this paper, we propose a semi-supervised sparse representation method to learn discriminative labels for the unlabeled data. In the method, we extract sparse representations of the data with SC and learn a data-driven classifier simultaneously. Considering the labels of data as label variables, then the variables can be obtained by projecting the sparse representations with the learned classifier. For better classification performance, we assume that the label variables should keep the same manifold structure as the original data. Moreover, the label variables for the data from the same classes should have similar representations, while the label variables for the data from diverse classes should deviate from each other. Taking these assumptions into consideration, we construct three graphs, i.e., the global manifold graph, the within-class graph and the between-class graph, to regularize the learning process of the label variables. Consequently, the features extracted by the proposed method not only keep the same manifold structure as the original data, but also are discriminative for classification. In this way, we unify the processes of sparse coding, data classification and the label variables regularization in one optimization problem. Experimental results demonstrate that the proposed method can achieve excellent performances for image classification problems with insufficient labeled data.

To summarize, the contributions of this paper can be listed as follows:
\begin{enumerate}
\item We propose a discriminative semi-supervised sparse representation method for image classification problems with insufficient labeled data. By making full use of the unlabeled data, the method is able to learn data labels that keep the same manifold structure as the original data and are discriminative for classification.
\item We construct the within-class graph and between-class graph in a different way to exploit the underlying the structure information included in the unlabeled data. By assigning additional weights to the unlabeled data within the neighborhood, the proposed method is able to learn discriminative features for classification.
\item We propose a kernel semi-supervised sparse representation method to efficiently classify images that cannot be linearly classified.
\item We develop two efficient algorithms to optimize the sparse coding problems for two proposed semi-supervised sparse representation algorithms respectively.
\end{enumerate}

The remaining part of the paper is organized as follows: Section \uppercase\expandafter{\romannumeral2} reviews some related works. Section \uppercase\expandafter{\romannumeral3} introduces the proposed semi-supervised sparse representation algorithm in detail. In Section \uppercase\expandafter{\romannumeral4}, we develop an efficient algorithm to solve the optimization problem. Then, the kernel semi-supervised sparse representation method and its corresponding optimization algorithm are presented in Section \uppercase\expandafter{\romannumeral5}. Experimental results on several databases are reported to evaluate the effectiveness of the proposed method in Section \uppercase\expandafter{\romannumeral6}. Finally, we draw conclusion and suggest possible future research directions in Section \uppercase\expandafter{\romannumeral7}.

\section{Related work}

In this section, we briefly review some previous works that are related to the proposed method.

In general, the semi-supervised learning (SSL) can be viewed as a class of supervised learning methods that can utilize the unlabeled data for training. Commonly, the labeled data are expensive to obtain, whereas the unlabeled data are relatively easy to acquire in reality. In the situations where there is only limited labeled data but massive unlabeled data, the SSL methods tend to achieve considerable improvement in accuracy compared with the related supervised methods. Therefore, SSL is more feasible for practical problems and of great interest to many researchers \cite{chapelle2006semi,cai2007semi,yang2012semi}.

To utilize the unlabeled data effectively, we usually need to make some assumptions to the underlying distributions of the data for the SSL methods. Amongst these, the smoothness assumption and manifold assumption are the two typical assumptions adopt by researchers. The smoothness assumption indicates that points which are close to each other tend to share the same labels \cite{chapelle2009semi}. Typical SSL methods based on the cluster assumption include Gaussian mixtures \cite{yu2012solving} and transductive support vector machine \cite{joachims2006transductive}. The manifold assumption supposes that data actually lie on a manifold with much lower dimensionality than that of the original data. If two points are within the neighborhood when projected onto the low dimensional manifold, then they are more likely to share the same labels. In other words, the data are assumed to have intrinsic manifold structure, along which the corresponding labels vary smoothly. It has been widely observed that better performance can be achieved with smooth solutions under this assumption \cite{gong2015deformed}. The Graph-based SSL (GSSL) algorithms are typical SSL methods based on this assumption and attract much attention recently \cite{wang2014semi,yu2006active,he2011nonnegative,zheng2013image}.

The GSSL methods describe the manifold structure of data with graphs constructed with the data points and the corresponding pairwise similarities. Then, a smoothness term will be designated to the graph to regularize the features or labels. Therefore, how to construct the graph plays a significant role in this kind of methods. To exploit the underlying geometrical structure of the data, Cai et al. \cite{zheng2011graph} proposed a graph regularized sparse coding method which meets the manifold assumption. He et al. \cite{he2011nonnegative} proposed a method to propagate the labels from the labeled data to the unlabeled data via sparse coding. However, the class labels of the labeled data are ignored during the sparse learning process. Therefore, the method still belongs to the unsupervised method. \cite{wang2014semi} and \cite{zheng2013image} proposed similar methods that utilize the manifold structure of the labeled and unlabeled data and the label constraint provided by the labeled data to learn labels for the unlabeled data. Therefore, the manifold constraint in these methods is weak and no discriminative property is enforced for the learned labels. As can be seen that successful methods should be able to utilize the labels of the training data and feature vectors of the labeled and unlabeled data simultaneously to learn discriminative features. In particular, the desirable features should be as smooth as the original data in the subspace while have discriminative properties for classification. Consequently, few existing sparse coding methods can meet the requirement of neighborhood smoothness and being discriminative among different labeled data at the same time.

To alleviate the above problems, a semi-supervised sparse representation method for image classification is proposed in this paper. In the method, we extract the sparse representations of images with SC and learn a data-driven classifier simultaneously. Particularly, we consider the labels of data as label variables. To obtain discriminative labels for the unlabeled data, we have assumptions: 1) the label variables should keep the same manifold structure as the original data; 2) the label variables for the same class should have similar representations, while the label variables for different classes should deviate from each other. Taking these assumption into consideration, three graphs, i.e., the global manifold graph, the within-class graph and the between-class graph, are constructed to regularize the learning process for the label variables. In this way, the features extracted by the proposed method not only keep the manifold structure of the original data, but also are discriminative for classification. Experimental results on challenging databases show that the proposed method are highly effective for image classification problems.

\section{Semi-supervised sparse representation with graph regularization}

In this section, we describe the proposed semi-supervised sparse representation algorithm in detail. The aim of the algorithm is to learn proper label representations for the unlabeled data. To make full use of the underlying information contained in the unlabeled data, the proposed method integrates the unlabeled data with the sparse coding and classifier learning processes. Moreover, the global manifold graph regularization together with the within-class and between-class graph regularizations are proposed to learn discriminative labels for the unlabeled data.

\subsection{Preliminaries}
Assume that we have labeled training data matrix $\bm{X}_{l}=[\bm{x}_{1},\bm{x}_{2},\dots,\bm{x}_{l}]\in \mathbb{R}^{d\times l}$, where $\bm{x}_{i}\in \mathbb{R}^{d},i=1,2,\dots,l$ is a labeled image with $d$ dimension and $l$ is the number of the training data. And we have unlabeled data $\bm{X}_{u}=[\bm{x}_{l+1},\bm{x}_{l+2},\dots,\bm{x}_{l+u}]\in \mathbb{R}^{d\times u}$, where $u$ is the number of the unlabeled data. Then we express all the data with a matrix $\bm{X}=[\bm{X}_{l},\bm{X}_{u}]\in \mathbb{R}^{d\times n}$, where $n=l+u$ is the total number of the data. We further assume that the data belong to $c$ different classes and the labels of the training data are defined by $\bm{Y}_{l}=[\bm{y}_{1},\bm{y}_{2},\dots,\bm{y}_{l}]\in \mathbb{R}^{c\times l}$. Each column $\bm{y}_{i}\in \mathbb{R}^{c},\ i=1,2,\cdots,l$ of $\bm{Y}_{l}$ is a binary vector in which the position of $``1"$ indicates the class of the corresponding data.

\subsection{Sparse representation}
Learning representative features of data for classification problems can be accomplished by solving the following sparse coding problem:
\begin{equation}\label{eq-01}
\begin{split}
<\bm{D},\bm{S}>&=\arg\min_{\bm{D},\bm{S}}\parallel \bm{X}-\bm{D}\bm{S}\parallel^{2}_{2}+\lambda\parallel \bm{S}\parallel_{0},\\
     &\textrm{s.t.}\ \parallel \bm{D}_{\cdot j}\parallel_{2}^{2}\leq1,\ j=1,2,\cdots,k
\end{split}
\end{equation}
where $\lambda$ is a non-negative regularization parameter for inducing sparsity of $\bm{S}$, $\parallel \bm{x}\parallel_{0}$ is the pseudo-$l_{0}$ norm that counts the number of non-zero elements in $\bm{x}$, and the inequalities regularize the norm of each column of $\bm{D}$ to be not bigger than $1$. $\bm{D}\in \mathbb{R}^{d\times k}$ is the dictionary with each column being an atom , while $\bm{S}$ is the sparse coefficient of the data $\bm{X}$ with most of its elements being zero.

According to the theory in \cite{candes2006stable,donoho2006compressed}, the pseudo-$l_{0}$ norm can be approximated by $l_{1}$ norm as long as the coefficient is sparse enough. Consequently, the above problem (\ref{eq-01}) can be relaxed to the following optimization problem:
\begin{equation}\label{eq-02}
\begin{split}
<\bm{D},\bm{S}>&=\arg\min_{\bm{D},\bm{S}}\parallel \bm{X}-\bm{D}\bm{S}\parallel^{2}_{2}+\lambda\parallel \bm{S}\parallel_{1},\\
     &\textrm{s.t.}\ \parallel \bm{D}_{\cdot j}\parallel_{2}^{2}\leq1,\ j=1,2,\cdots,k
\end{split}
\end{equation}
where $\parallel \bm{x}\parallel_{1}$ is the $l_{1}$ norm which sums the absolute values of all elements in $\bm{x}$.

\subsection{Construction of linear classifier for classification}
After the sparse representations of the original data are obtained through the sparse coding process, they are input to a classifier to complete the classification tasks. Popular classifiers include $K$-nearest neighbors ($K$-NN), support vector machine (SVM) and so on. However, these classifiers are not specifically designed for concrete classification problems. Therefore, developing a data-driven classifier for a specific classification problem is more desirable. Suppose that the labels of data $\bm{X}$ could be approximated from the corresponding sparse representatives $\bm{S}\in \mathbb{R}^{k\times n}$:
\begin{equation}\label{eq-03}
\bm{H}=\bm{W}\bm{S},\ \textrm{s.t.}\ \parallel \bm{W}_{\cdot m}\parallel_{2}^{2}\leq1,\ m=1,2,\cdots,k
\end{equation}
where $\bm{W}\in \mathbb{R}^{c\times k}$ is the classifier matrix and its columns are constrained by the inequalities, and $\bm{H}=[\bm{h}_{1},\bm{h}_{2},\cdots,\bm{h}_{n}]\in \mathbb{R}^{c\times n}$ is the predicted label matrix. Each column $\bm{h}_{i}\in \mathbb{R}^{c},\ i=1,2,\cdots,n$ indicates the predicted label information for the corresponding data. The equation (\ref{eq-03}) implies that the labels of the data $\bm{X}$ can be predicted via a linear transformation of the sparse coefficient.

To obtain the class number of each data, maximum operation is performed to each column of the matrix $\bm{H}$, i.e.,
\[c_{j}=\arg\max_{i\in\{1,2,\cdots,c\}}\bm{H}_{ij},\ j=1,2,\cdots,n\]
where $c_{j}$ is the class of the $j$-th data.

Combining (\ref{eq-02}) with (\ref{eq-03}), we have the following optimization problem:
\begin{equation}\label{eq-04}
\begin{split}
<\bm{D},\bm{S},\bm{W},\bm{H}>&=\arg\min_{\bm{D},\bm{S},\bm{W},\bm{H}}\parallel \bm{X}-\bm{D}\bm{S}\parallel^{2}_{2}+\\
       &\lambda\parallel \bm{S}\parallel_{1}+\alpha\parallel \bm{H}-\bm{W}\bm{S}\parallel_{2}^{2}\\
       &+\gamma \textrm{Tr}((\bm{H}-\bm{F})\bm{U}(\bm{H}-\bm{F})^{T})\\
       &\textrm{s.t.}\ \parallel \bm{D}_{\cdot j}\parallel_{2}^{2}\leq1,\ j=1,2,\cdots,k\\
       &\ \ \ \ \ \parallel \bm{W}_{\cdot m}\parallel_{2}^{2}\leq1,\ m=1,2,\cdots,k
\end{split}
\end{equation}
where $\alpha$ is the parameter that controls the weight of the classification error, $\bm{U}\in \mathbb{R}^{n\times n}$ is a diagonal matrix whose first $l$ diagonal elements are ones and others are zeros, and $\bm{F}\in \mathbb{R}^{c\times n}$ is a matrix with the first $l$ columns being equal to $\bm{Y}_{l}$ while others being zeros. The last term in the above equation enforces the predicted labels $\bm{H}$ to be consistent with the available training labels.

\subsection{Construction of the discriminative label prediction}
Though (\ref{eq-04}) is very effective for a classification problem, however, it does not enforce discriminative properties for better classification performance. Therefore, we propose a discriminative method to further enhance the performance of (\ref{eq-04}). It is natural to require that the data belonging to the same classes have similar label vectors, while the data from diverse classes have dissimilar label vectors. As aforementioned, each column of the predicted label matrix $\bm{H}$ indicates the label information. Suppose that each column $\bm{h}_{i},\ i=1,2,\cdots,n$ of the label matrix $\bm{H}$ is regarded as a point in the space $\mathbb{R}^{c}$, then for all the point pairs, they should have large between-class distances and have small within-class distances. At the same time, the label vectors should also keep the same manifold structure as the original data $\bm{X}$.

To achieve this, we propose to construct three graphs $G_{w}$, $G_{b}$ and $G_{g}$. $G_{w}$ is the within-class graph which defines the affinity of the data from the same classes, $G_{b}$ is the between-class graph which defines the affinity of the data from diverse classes, while the $G_{g}$ is the global manifold graph which describes the global manifold structure of the whole data. The nodes in the graphs are the data $\{\bm{x}_{i}\}^{n}_{i=1}$ and their edges are defined by affinity matrixes $\bm{A}^{w}$, $\bm{A}^{b}$ and $\bm{A}^{g}$ respectively. An affinity matrix measures the similarity between any two nodes in the graph. For point $\bm{x}_{i}$, we denote $\bm{N}_{k}(\bm{x}_{i})$ as the $k$ nearest neighbors set with Euclidian metric. Before introducing the three graphs, we define the affinity matrix:
\begin{eqnarray}
\bm{A}_{ij}=\left\{\begin{array}{ll}
  1,&\bm{x}_{i}\in \bm{N}_{k}(\bm{x}_{j})\ \textrm{or}\ \bm{x}_{j}\in \bm{N}_{k}(\bm{x}_{i})\\
  0,&\textrm{others}\\
\end{array}\right.\nonumber
\end{eqnarray}

\subsubsection{Construction of the within-class graph $G_{w}$}
The weight between any two points $\bm{x}_{i}$ and $\bm{x}_{j}$ in the graph $G_{w}$ will be given with a affinity matrix $\bm{A}^{w}$. The affinity matrix $\bm{A}^{w}$ is defined as:
\begin{eqnarray}
\bm{A}^{w}_{ij}=\left\{\begin{array}{ll}
     \bm{A}_{ij}/n_{\bm{y}_{i}}+\beta_{w} \bm{A}_{ij},&\bm{y}_{i}=\bm{y}_{j}\\
     \beta_{w} \bm{A}_{ij},&\bm{x}_{j}\in \bm{N}_{k}(\bm{x}_{i})\ \textrm{or}\ \bm{x}_{i}\in \bm{N}_{k}(\bm{x}_{j})\\
     0,&\textrm{others}
     \end{array}\right.\nonumber
\end{eqnarray}
where $n_{\bm{y}_{i}}$ is the number of points with the same label $\bm{y}_{i}$ and $\beta_{w}\geq0$ is a constant. As the points within the neighborhood are supposed to have same labels, a small weight should be assigned to the unlabeled points in the neighborhood. To ensure that the labeled data from the same classes in the neighborhood always have relatively larger weights than those of the unlabeled ones, extra weights are added to the labeled points at the same time. In this way, the affinity matrix $\bm{A}^{w}$ defines the similarities between points from both labeled and unlabeled data. Consequently, underlying information lies in the unlabeled data can be fully exploited. Note that $\bm{A}^{w}$ will degenerate to the within-class affinity matrix proposed in \cite{sugiyama2006local} when $\beta_{w}=0$, which is only defined for labeled data.

With the affinity matrix $\bm{A}^{w}$, the summation of the within-class distance of the label vectors $\{\bm{h}_{i}\}_{i=1}^{n}$ can be computed as:
\begin{eqnarray}\label{eq-05}
\bm{Q}_{w}&=&\sum_{ij}\parallel \bm{h}_{i}-\bm{h}_{j}\parallel^{2}\bm{A}^{w}_{ij}\nonumber\\
        &=&2\sum_{i}(\bm{h}_{i}^{T}\bm{D}^{w}_{ii}\bm{h}_{i})-2\sum_{ij}(\bm{h}_{i}^{T}\bm{A}^{w}_{ij}\bm{h}_{j})\\
        &=&2\textrm{Tr}(\bm{H}(\bm{D}^{w}-\bm{A}^{w})\bm{H}^{T})\nonumber\\
        &=&2\textrm{Tr}(\bm{H}\bm{L}^{w}\bm{H}^{T})\nonumber
\end{eqnarray}
where $\bm{D}^{w}$ is a diagonal matrix whose diagonal entries are the summations of corresponding rows of $\bm{A}^{w}$, i.e., $\bm{D}^{w}_{ii}=\sum_{j}\bm{A}^{w}_{ij}$, and $\bm{L}^{w}=\bm{D}^{w}-\bm{A}^{w}$ is the graph Laplacian matrix.

\subsubsection{Construction of the between-class graph $G_{b}$}
Similarly, the weight between any two points $\bm{x}_{i}$ and $\bm{x}_{j}$ in the graph $G_{b}$ will be given by an affinity matrix $\bm{A}^{b}$. The affinity matrix $\bm{A}^{b}$ can be defined as:
\begin{eqnarray}
\bm{A}^{b}_{ij}=\left\{\begin{array}{ll}
     \bm{A}_{ij}(1/n-1/n_{\bm{y}_{i}})-\beta_{b} \bm{A}_{ij},&\bm{y}_{i}=\bm{y}_{j}\\
     -\beta_{b} \bm{A}_{ij},&\bm{x}_{j}\in \bm{N}_{k}(\bm{x}_{i})\\
     &\textrm{or}\ \bm{x}_{i}\in \bm{N}_{k}(\bm{x}_{j})\\
     \bm{A}_{ij}/n,&\bm{y}_{i}\neq \bm{y}_{j}\\
     0,&\textrm{others}
     \end{array}\right.\nonumber
\end{eqnarray}
where $\beta_{b}\geq0$ is a constant that weights the points in the neighborhood.

Then, we can compute the summation of the between-class distance of the label vectors $\{\bm{h}_{i}\}_{1}^{n}$ as:
\begin{eqnarray}\label{eq-06}
\bm{Q}_{b}&=&\sum_{ij}\parallel \bm{h}_{i}-\bm{h}_{j}\parallel^{2}\bm{A}^{b}_{ij}\nonumber\\
        &=&2\sum_{i}(\bm{h}_{i}^{T}\bm{D}^{b}_{ii}\bm{h}_{i})-2\sum_{ij}(\bm{h}_{i}^{T}\bm{A}^{b}_{ij}\bm{h}_{j})\\
        &=&2\textrm{Tr}(\bm{H}(\bm{D}^{b}-\bm{A}^{b})\bm{H}^{T})\nonumber\\
        &=&2\textrm{Tr}(\bm{H}\bm{L}^{b}\bm{H}^{T})\nonumber
\end{eqnarray}
where $\bm{D}^{b}$ is a diagonal matrix whose diagonal entries are the summations of the corresponding rows of $\bm{A}^{b}$, i.e., $\bm{D}^{b}_{ii}=\sum_{j}\bm{A}^{b}_{ij}$, and $\bm{L}^{b}=\bm{D}^{b}-\bm{A}^{b}$ is the graph Laplacian matrix.

\subsubsection{Construction of the global manifold graph $G_{g}$}
Besides the constraints of the within-class and between-class distances, we should enfore more constraints to each point to further shrink the distances between similar points. Here we employ the similarity propagation constraint proposed in \cite{zhang2014hyperspectral} to restrict the points from the same classes and the ones from diverse classes simultaneously. The similarity constraint can be expressed with an optimal intrinsic similarity matrix $\bm{P}$ that measures the similarities between all points by propagating a strong similarity (defined with label information) to all points with a weak similarity.

As we know, the affinity matrix $\bm{A}$ holds weak similarities for the points as no supervised information from the labeled data is enforced. To include the supervised information, we construct a strong similarity matrix $\bm{G}\in \mathbb{R}^{n\times n}$:
\begin{eqnarray}
\bm{G}_{ij}=\left\{\begin{array}{ll}
  1,&\bm{x}_{j}\in \bm{N}_{k}(\bm{x}_{i})\ \textrm{and}\ \bm{y}_{i}=\bm{y}_{j}\\
  0,&\textrm{others}\\
\end{array}\right.\nonumber
\end{eqnarray}
and let $\bm{G}_{ii}=0$ for $i=1,2,\cdots,n$.

Then we initialize the matrix $\bm{P}$ by setting $\bm{P}^{(0)}=\bm{G}$ and $\bm{P}^{(0)}_{ii}=1$ for $i=1,2,\cdots,n$. The elements with $\bm{P}^{(0)}_{ij}=1$ are regarded as original positive energies which will be propagated to other elements with $\bm{P}^{(0)}_{ij}=0$, following the path built in the weak similarity matrix $\bm{A}$. We formulate the criterion of the similarity propagation as \cite{liu2010constrained}:
\begin{equation}\label{eq-07}
\bm{P}^{(t+1)}_{i\cdot}=(1-\gamma)\bm{P}^{(0)}_{i\cdot}+\gamma\frac{\sum_{j}\bm{A}_{ij}\bm{P}^{(t)}_{j\cdot}}{\sum_{j}\bm{A}_{ij}}
\end{equation}
where $\bm{P}^{(t)}_{i\cdot}$ is the $i$-th row of matrix $\bm{P}$ at the $t$-th step and $0<\gamma<1$ is a parameter indicating the relative amount of the information from its neighbors and initial supervised information \cite{zhang2014hyperspectral}. The equation (\ref{eq-07}) can be further written as:
\[\bm{P}^{(t+1)}=(1-\gamma)\bm{P}^{(0)}+\gamma \bm{T}\bm{P}^{(t)}\]
where $\bm{T}=\bm{D}^{-1}\bm{A}$ is the well-known transition probability matrix in the Markov random walk models and $\bm{D}$ is a diagonal matrix with its diagonal elements $\bm{D}_{ii}=\sum_{j}\bm{A}_{ij}$, for $i=1,2,\cdots,n$.

As $0<\gamma<1$ and the eigenvalues of $\bm{T}$ are in $[-1,1]$, the sequence $\{\bm{P}^{(t)}\}$ converges to a limit value \cite{zhou2004learning}:
\[\bm{P}^{*}=\lim_{t\rightarrow\infty}\bm{P}^{(t)}=(1-\gamma)(\bm{I}-\gamma \bm{T})^{-1}\bm{P}^{(0)}\]
It should be noted that $(1-\gamma \bm{T})^{-1}$ is actually a graph or diffusion kernel \cite{kandola2002learning}. Finally, the expected $\bm{P}$ can be obtained by symmetrizing $\bm{P}^{*}$ and removing the tiny values, i.e.,
\begin{equation}\label{eq-08}
\bm{P}=(\frac{\bm{P}^{*}+\bm{P}^{*\bm{T}}}{2})_{\geq\delta}
\end{equation}
In equation (\ref{eq-08}), we set the values smaller than $\delta$ to be $0$.

The similarity matrix $\bm{P}$ reflects the global manifold structure of the whole data $\bm{X}$ and links the similar points and dissimilar ones simultaneously. We hope that the predicted label matrix $\bm{H}$ can hold the same manifold structure in the space $\mathbb{R}^{c}$ as the original data. Therefore, with the similarity matrix $\bm{P}$, we sum the pair-wise distances of the vectors $\{\bm{h}_{i}\}_{i=1}^{n}$ as:
\begin{eqnarray}\label{eq-09}
\bm{Q}_{g}&=&\sum_{ij}\parallel \bm{h}_{i}-\bm{h}_{j}\parallel^{2}\bm{P}_{ij}\nonumber\\
     &=&2\sum_{i}(\bm{h}_{i}^{T}\bm{D}^{g}_{ii}\bm{h}_{i})-2\sum_{ij}(\bm{h}_{i}^{T}\bm{P}_{ij}\bm{h}_{j})\\
     &=&2\textrm{Tr}(\bm{H}(\bm{D}^{g}-\bm{P})\bm{H}^{T})\nonumber\\
     &=&2\textrm{Tr}(\bm{H}\bm{L}^{g}\bm{H}^{T})\nonumber
\end{eqnarray}
where $\bm{D}^{g}$ is a diagonal matrix whose diagonal elements are the summations of the corresponding rows of $\bm{P}$, i.e., $\bm{D}^{g}_{ii}=\sum_{j}\bm{P}_{ij}$. And $\bm{L}^{g}=\bm{D}^{g}-\bm{P}$ is the graph Laplacian matrix.

To make the predicted label vectors $\{\bm{h}_{i}\}_{i=1}^{n}$ have small within-class distance and large between-class distance while keeping the same global manifold structure as the original data $\bm{X}$, we combine equation (\ref{eq-04}) with equations (\ref{eq-05}), (\ref{eq-06}) and (\ref{eq-09}) to obtain the following unified optimization problem:
\begin{equation}\label{eq-10}
\begin{split}
<\bm{D},\bm{S},\bm{W},\bm{H}>&=\arg\min_{\bm{D},\bm{S},\bm{W},\bm{H}}\parallel \bm{X}-\bm{D}\bm{S}\parallel^{2}_{2}+\lambda\parallel \bm{S}\parallel_{1}\\
       &+\alpha\parallel \bm{H}-\bm{W}\bm{S}\parallel_{2}^{2}+\beta_{1}\textrm{Tr}(\bm{H}\bm{L}^{g}\bm{H}^{T})\\
       &+\beta_{2}\textrm{Tr}(\bm{H}\bm{L}^{w}\bm{H}^{T})-\beta_{3}\textrm{Tr}(\bm{H}\bm{L}^{b}\bm{H}^{T})\\
       &+\gamma \textrm{Tr}((\bm{H}-\bm{F})\bm{U}(\bm{H}-\bm{F})^{T})\\
       &\textrm{s.t.}\ \parallel \bm{D}_{\cdot j}\parallel_{2}^{2}\leq1,\ j=1,2,\cdots,k\\
       &\ \ \ \ \ \parallel \bm{W}_{\cdot m}\parallel_{2}^{2}\leq1,\ m=1,2,\cdots,k
\end{split}
\end{equation}
The fourth term on the right side of the equation (\ref{eq-10}) enforces the predicted label vectors $\{\bm{h}_{i}\}_{i=1}^{n}$ to have the same manifold structure as the original data, the fifth term enforces the vectors to have small within-class distance, the sixth term enforces large between-class distance, while the last term makes the predicted labels be consistent with the available training labels.

We rewrite the equation (\ref{eq-10}) in a more compact form:
\begin{equation}\label{eq-11}
\begin{split}
<\bm{D},\bm{S},\bm{W},\bm{H}>&=\arg\min_{\bm{D},\bm{S},\bm{W},\bm{H}}\parallel \bm{X}-\bm{D}\bm{S}\parallel^{2}_{2}+\lambda\parallel \bm{S}\parallel_{1}\\
       &+\alpha\parallel \bm{H}-\bm{W}\bm{S}\parallel_{2}^{2}+\textrm{Tr}(\bm{H}\bm{L}\bm{H}^{T})\\
       &+\gamma \textrm{Tr}((\bm{H}-\bm{F})\bm{U}(\bm{H}-\bm{F})^{T})\\
       &\textrm{s.t.}\ \parallel \bm{D}_{\cdot j}\parallel_{2}^{2}\leq1,\ j=1,2,\cdots,k\\
       &\ \ \ \ \ \parallel \bm{W}_{\cdot m}\parallel_{2}^{2}\leq1,\ m=1,2,\cdots,k
\end{split}
\end{equation}
where $\bm{L}=\beta_{1}\bm{L}^{g}+\beta_{2}\bm{L}^{w}-\beta_{3}\bm{L}^{b}$.

\section{Optimization methods}
In this section, we will propose an efficient algorithm to solve the optimization problem (\ref{eq-11}). As can be seen that there are four variables in the problem, i.e., the dictionary $\bm{D}$, the sparse coefficient $\bm{S}$, the classifier matrix $\bm{W}$ and the predicted label matrix $\bm{H}$. We will solve the problem in an alternating way, i.e., update one variable each time while fixing the others.

\subsection{Initialization}
\label{sec:initial}
We need to initialize the variables: the dictionary $\bm{D}_{0}$, the sparse coefficient $\bm{S}_{0}$, the classifier matrix $\bm{W}_{0}$ and the predicted label matrix $\bm{H}_{0}$. The dictionary $\bm{D}_{0}$ and sparse coefficient $\bm{S}_{0}$ can be initialized through solving the sparse coding problem (\ref{eq-02}) with the Lagrange Dual algorithm \cite{lee2006efficient} and the alternating direction method of multipliers (ADMM) \cite{boyd2011distributed} respectively.

To initialize $\bm{W}_{0}$, we utilize the multivariate ridge regression model \cite{golub1999tikhonov}, with the quadratic loss and $l_{2}$ norm regularization. It can be expressed as:
\[\bm{W}_{0}=\arg\min_{\bm{W}}\alpha\parallel \bm{H}-\bm{W}\bm{S}\parallel_{2}^{2}+\mu\parallel \bm{W}\parallel_{2}^{2}\]
where $\mu$ is a parameter to regularize $\bm{W}$. The above problem has the following solution:
\[\bm{W}_{0}=\alpha \bm{H}\bm{S}^{T}(\alpha \bm{S}\bm{S}^{T}+\mu \bm{I})^{-1}\]

To initialize $\bm{H}_{0}$, we let the first $l$ columns of $\bm{H}_{0}$ contain the true label information, i.e., $\bm{H}_{0,\cdot i}=\bm{y}_{i}$, $i=1,2,\cdots,l$, while the other values are generated randomly.

\subsection{Optimization of the dictionary $\bm{D}$ and classifier matrix $\bm{W}$}
We first discuss the optimization of the dictionary $\bm{D}$ and the classifier matrix $\bm{W}$. The $\bm{D}$ and $\bm{W}$ can be optimized simultaneously as they can be concatenated to form a generalized dictionary. Fixing $\bm{S}$ and $\bm{H}$ and removing other unrelated items, we have the following optimization problem:
\begin{equation}\label{eq-12}
\begin{split}
<\bm{D},\bm{W}>&=\arg\min_{\bm{D},\bm{W}}\parallel \bm{X}-\bm{D}\bm{S}\parallel^{2}_{2}+\alpha\parallel \bm{H}-\bm{W}\bm{S}\parallel_{2}^{2}\\
     &\textrm{s.t.}\ \parallel \bm{D}_{\cdot j}\parallel_{2}^{2}\leq1,\ j=1,2,\cdots,k\\
     &\ \ \ \ \ \parallel \bm{W}_{\cdot m}\parallel_{2}^{2}\leq1,\ m=1,2,\cdots,k
\end{split}
\end{equation}
We define $\tilde{\bm{X}}=\begin{bmatrix}\bm{X}\\\sqrt{\alpha}\bm{H}\end{bmatrix}$ as an extended data matrix and $\tilde{\bm{D}}=\begin{bmatrix}\bm{D}\\\sqrt{\alpha}\bm{W}\end{bmatrix}$ as an extended dictionary matrix. Consequently, the inequality constraints become $\parallel\tilde{\bm{D}}_{\cdot j}\parallel_{2}^{2}\leq1+\sqrt\alpha,\ j=1,2,\cdots,k$. Then the equation (\ref{eq-12}) can be rewritten as:
\begin{equation}\label{eq-13}
\begin{split}
<\tilde{\bm{D}}>&=\arg\min_{\tilde{\bm{D}}}\parallel \tilde{\bm{X}}-\tilde{\bm{D}}\bm{S}\parallel^{2}_{2}\\
           &\textrm{s.t.}\ \parallel\tilde{\bm{D}}_{\cdot j}\parallel_{2}^{2}\leq1+\sqrt\alpha,\ j=1,2,\cdots,k
\end{split}
\end{equation}

The problem (\ref{eq-13}) could be solved by the Lagrange Dual algorithm proposed in \cite{lee2006efficient}. Then the desired dictionary $\bm{D}$ and classifier matrix $\bm{W}$ can be obtained through $\tilde{\bm{D}}$ as:
\begin{equation}\label{eq-15}
\begin{split}
\bm{D}&=\tilde{\bm{D}}_{1:d,\cdot}\\
\bm{W}&=\frac{1}{\sqrt{\alpha}}\tilde{\bm{D}}_{d+1:d+c,\cdot}
\end{split}
\end{equation}
where $\tilde{\bm{D}}_{i,\cdot},\ i=1,2,\cdots,d+c$ is the $i$-th row of $\tilde{\bm{D}}$.

\subsection{Optimization of the sparse coefficient $\bm{S}$}
Fixing $\bm{D}$, $\bm{W}$, $\bm{H}$ and removing other irrelevant terms, the optimization problem for $\bm{S}$ is:
\begin{equation}\label{eq-16-0}
<\bm{S}>=\arg\min_{\bm{S}}\parallel \tilde{\bm{X}}-\tilde{\bm{D}}\bm{S}\parallel^{2}_{2}+\lambda\parallel \bm{S}\parallel_{1}
\end{equation}
The problem (\ref{eq-16-0}) is a standard sparse coding problem which can be solved with the ADMM algorithm \cite{boyd2011distributed}.

\begin{algorithm}[!t]
\caption{Semi-supervised sparse representation with graph regularization algorithm (SSRGR)}
\label{alg1}
\algsetup{linenodelimiter=.}
\begin{algorithmic}[1]
\REQUIRE The image data $\bm{X}$, training label matrix $\bm{F}$, and label regularization matrix $\bm{U}$.
\STATE Initialize $\bm{D}_{0}$, $\bm{S}_{0}$, $\bm{W}_{0}$ and $\bm{H}_{0}$ according to the description in the Subsection \ref{sec:initial}.
\FOR{$i=1$ to $J_{1}$}
\STATE Compute the extended data $\tilde{\bm{X}}=[\bm{X},\sqrt{\alpha}\bm{H}^{T}]^{T}$ and dictionary $\tilde{\bm{D}}=[\bm{D},\sqrt{\alpha}\bm{W}]^{T}$;
\STATE Update the dictionary $\bm{D}$ and the classifier matrix $\bm{W}$ according to (\ref{eq-15}) ;
\STATE Update the sparse coefficient $\bm{S}$ with ADMM \cite{boyd2011distributed};
\STATE Update the predicted label matrix $\bm{H}$ according to (\ref{eq-16}).
\ENDFOR
\ENSURE The dictionary $\bm{D}$, sparse coefficient $\bm{S}$, classifier matrix $\bm{W}$ and predicted label matrix $\bm{H}$.
\end{algorithmic}
\end{algorithm}

\subsection{Optimization of the predicted label matrix $\bm{H}$}
Fixing $\bm{D}$, $\bm{S}$ and $\bm{W}$, we can update $\bm{H}$ via the following optimization problem:
\begin{equation}\label{eq-16}
\begin{split}
<\bm{H}>&=\arg\min_{\bm{H}}\alpha\parallel \bm{H}-\bm{W}\bm{S}\parallel_{2}^{2}+\textrm{Tr}(\bm{H}\bm{L}\bm{H}^{T})+\\
   &\gamma \textrm{Tr}((\bm{H}-\bm{F})\bm{U}(\bm{H}-\bm{F})^{T})
\end{split}
\end{equation}
The solution to the problem (\ref{eq-16}) is:
\[\bm{H}=(\alpha \bm{W}\bm{S}+\gamma \bm{F}\bm{U})(\alpha \bm{I}+(\bm{L}+\bm{L}^{T})/2+\gamma \bm{U})^{-1}\]

Finally, we summarize the algorithm to solve the optimization problem (\ref{eq-11}) with Algorithm \ref{alg1}.

\section{Kernel semi-supervised sparse representation with graph regularization and its optimization methods}

The proposed SSRGR algorithm tries to learn a data-driven classifier for image classification problems. However, it may fail to classify data generated with nonlinear structures. One feasible solution to this situation is to project the data into a higher dimensional space with kernel functions and classify it with linearly in the projected space. In the following, we introduce the kernel version of the SSRGR algorithm, which is called the KSSRGR algorithm.

\subsection{Kernel semi-supervised sparse representation with graph regularization}

Suppose that there is a nonlinear mapping function $\bm{\phi}(x): \mathbb{R}^{d}\rightarrow \mathbb{R}^{d^{'}}$, where $d\leq d^{'}$ and $d^{'}$ is the dimensionality of the projected high dimensional space, then we have $\bm{\phi}(\bm{X})=[\bm{\phi}(\bm{x}_{1}),\bm{\phi}(\bm{x}_{2}),\cdots,\bm{\phi}(\bm{x}_{n})]$. However, we do not define the nonlinear mapping function $\bm{\phi}(x)$ explicitly, but implicitly express it via a kernel function which is defined as the inner production of two mapping functions, i.e., $\bm{K}(\bm{x}_{i},\bm{x}_{j})=\bm{\phi}(\bm{x}_{i})^{T}\bm{\phi}(\bm{x}_{j})$. Therefore, given the data matrix $\bm{X}$ and the nonlinear mapping matrix $\bm{\phi}(\bm{X})$, the kernel matrix can be defined as:
\begin{equation}
\begin{split}
\bm{K}&=\bm{\phi}(\bm{X})^{T}\bm{\phi}(\bm{X})\\
\bm{K}_{ij}&=\bm{\phi}(\bm{x}_{i})^{T}\bm{\phi}(\bm{x}_{j})=\bm{K}(\bm{x}_{i},\bm{x}_{j})\nonumber
\end{split}
\end{equation}

The commonly used kernel function is $\bm{K}(x,y)=e^{-\frac{\parallel x-y\parallel_{2}^{2}}{\sigma^{2}}}$. We employ it in this paper to compute the similarity between two points in the projected high dimensional space.

Furthermore, we assume that the columns of the dictionary $\bm{D}$ can be represented by the linear combination of the columns of the $\bm{\phi}(\bm{X})$ \cite{van2013design}, i.e.,
\[\bm{D}_{\cdot j}=\sum_{i=1}^{n}\bm{B}_{ij}\phi(\bm{x}_{i}),\]
where $\bm{B}_{ij}$ is the $(i,j)$-th element of the matrix $\bm{B}\in \mathbb{R}^{n\times k}$. Thus, we have $\bm{D}=\bm{\phi}(\bm{X})\bm{B}$. Substituting $\bm{\phi}(\bm{X})$ and $\bm{D}$ to the equation (\ref{eq-02}), then the sparse coding problem becomes:
\begin{equation}\label{eq-18}
\begin{split}
<\bm{B},\bm{S}>&=\arg\min_{\bm{B},\bm{S}}\parallel \bm{\phi}(\bm{X})-\bm{\phi}(\bm{X})\bm{B}\bm{S}\parallel^{2}_{2}+\lambda\parallel \bm{S}\parallel_{1}\\
     &=\arg\min_{\bm{B},\bm{S}}\textrm{Tr}((\bm{I}-\bm{B}\bm{S})^{T}\bm{\phi}(\bm{X})^{T}\bm{\phi}(\bm{X})(\bm{I}-\bm{B}\bm{S}))\\
     &\ \ \ +\lambda\parallel \bm{S}\parallel_{1}\\
     &=\arg\min_{\bm{B},\bm{S}}\textrm{Tr}((\bm{I}-\bm{B}\bm{S})^{T}\bm{K}(\bm{I}-\bm{B}\bm{S}))+\lambda\parallel \bm{S}\parallel_{1}\\
     &\textrm{s.t.}\ ||\bm{\phi}(\bm{X})\bm{B}_{\cdot j}||_{2}^{2}\leq1,\ j=1,2,\cdots,k
\end{split}
\end{equation}
where the inequalities indicate the constraints to the columns of the dictionary. The left side of the inequalities can be formulated as:
\begin{equation}
\begin{split}
\parallel \bm{\phi(X)}\bm{B}_{\cdot j}\parallel^{2}_{2}&=(\bm{\phi}(\bm{X})\bm{B}_{\cdot j})^{T}(\bm{\phi}(\bm{X})\bm{B}_{\cdot j})\\
                                             &=\bm{B}_{\cdot j}^{T}\bm{\phi}(\bm{X})^{T}\bm{\phi}(\bm{X})\bm{B}_{\cdot j}\\
                                             &=\bm{B}_{\cdot j}^{T}\bm{K}\bm{B}_{\cdot j}
\end{split}
\end{equation}
Thus, it is unnecessary to know the nonlinear mapping $\bm{\phi}(x)$ exactly, while $\bm{K}$ can be computed with the chosen kernel function.

Next, we update the definition of the graphs $G_{g}$, $G_{w}$ and $G_{b}$ respectively. As each data point $\bm{x}_{i}\in \mathbb{R}^{d}$ is mapped to a point $\bm{\phi}(\bm{x}_{i})\in \mathbb{R}^{d^{'}}$ in the high dimensional space, the distances among data should also be computed with a new metric. Let the distance between two points $\bm{\phi}(\bm{x}_{i})$ and $\bm{\phi}(\bm{x}_{j})$ be:
\[\begin{split}\tilde{d}(\bm{x}_{i},\bm{x}_{j})&=\parallel\bm{\phi}(\bm{x}_{i})-\bm{\phi}(\bm{x}_{j})\parallel_{2}^{2}\\
  &=\bm{\phi}(\bm{x}_{i})^{T}\bm{\phi}(\bm{x}_{i})-2\bm{\phi}(\bm{x}_{i})^{T}\bm{\phi}(\bm{x}_{j})+\bm{\phi}(\bm{x}_{i})^{T}\bm{\phi}(\bm{x}_{j})\\
  &=\bm{K}(\bm{x}_{i},\bm{x}_{i})-2\bm{K}(\bm{x}_{i},\bm{x}_{j})+\bm{K}(\bm{x}_{j},\bm{x}_{j})
\end{split}\]

We denote the $k$ nearest neighbors of the point $\bm{\phi}(\bm{x}_{i})$ as $\tilde{\bm{N}}_{k}(\bm{\phi}(\bm{x}_{i}))$. In exactly the same way, one can construct the affinity matrix $\tilde{\bm{A}}$, the affinity matrix $\tilde{\bm{A}}^w$ and the corresponding graph Laplacian $\tilde{\bm{L}}^w$ for the within-class graph, the affinity matrix $\tilde{\bm{A}}^b$ and the corresponding graph Laplacian $\tilde{\bm{L}}^b$ for the between-class graph, and the affinity matrix $\tilde{\bm{P}}$ and the corresponding graph Laplacian $\tilde{\bm{L}}^g$ for the global graph of the whole data.  Now the optimization problem equivalent to (\ref{eq-10}) involving the kernel $\bm{K}$ becomes:
\begin{equation}\label{eq-24}
\begin{split}
<\bm{B},\bm{S},\bm{W},\bm{H}>&=\arg\min_{\bm{B},\bm{S},\bm{W},\bm{H}}\textrm{Tr}((\bm{I}-\bm{B}\bm{S})^{T}\bm{K}(\bm{I}-\bm{B}\bm{S}))+\\
     &\lambda\parallel \bm{S}\parallel_{1}+\alpha\parallel \bm{H}-\bm{W}\bm{S}\parallel_{2}^{2}+\beta_{1}\textrm{Tr}(\bm{H}\tilde{\bm{L}}^{g}\bm{H}^{T})\\
     &+\beta_{2}\textrm{Tr}(\bm{H}\tilde{\bm{L}}^{w}\bm{H}^{T})-\beta_{3}\textrm{Tr}(\bm{H}\tilde{\bm{L}}^{b}\bm{H}^{T})\\
     &+\gamma \textrm{Tr}((\bm{H}-\bm{F})\bm{U}(\bm{H}-\bm{F})^{T})\\
     &\textrm{s.t.}\ \parallel\bm{\phi}(\bm{X})\bm{B}_{\cdot j}\parallel_{2}^{2}\leq1,\ j=1,2,\cdots,k\\
     &\ \ \ \ \ \parallel \bm{W}_{\cdot m}\parallel_{2}^{2}\leq1,\ m=1,2,\cdots,k
\end{split}
\end{equation}
which can be further written in a more compact form:
\begin{equation}\label{eq-25}
\begin{split}
<\bm{B},\bm{S},\bm{W},\bm{H}>&=\arg\min_{\bm{B},\bm{S},\bm{W},\bm{H}}\textrm{Tr}((\bm{I}-\bm{B}\bm{S})^{T}\bm{K}(\bm{I}-\bm{B}\bm{S}))\\
     &+\lambda\parallel \bm{S}\parallel_{1}+\alpha\parallel \bm{H}-\bm{W}\bm{S}\parallel_{2}^{2}+\textrm{Tr}(\bm{H}\tilde{\bm{L}}\bm{H}^{T})\\
     &+\gamma \textrm{Tr}((\bm{H}-\bm{F})\bm{U}(\bm{H}-\bm{F})^{T})\\
     &\textrm{s.t.}\ \parallel\bm{\phi}(\bm{X})\bm{B}_{\cdot j}\parallel_{2}^{2}\leq1,\ j=1,2,\cdots,k\\
     &\ \ \ \ \ \parallel \bm{W}_{\cdot m}\parallel_{2}^{2}\leq1,\ m=1,2,\cdots,k
\end{split}
\end{equation}
where $\tilde{\bm{L}}=\beta_{1}\tilde{\bm{L}}^{g}+\beta_{2}\tilde{\bm{L}}^{w}-\beta_{3}\tilde{\bm{L}}^{b}$.

We optimize the above problem (\ref{eq-25}) similarly as the Algorithm \ref{alg1} with an alternating method which is described in the following.

\subsection{Optimization of the dictionary associated matrix $\bm{B}$}
As discussed above, the dictionary D in the kernel algorithm can be expressed as $\bm{D}=\bm{\phi}(\bm{X})\bm{B}$. Consequently, the matrix $\bm{D}$ could be obtained by optimizing $\bm{B}$ in the equation (\ref{eq-25}). Fixing other variables, the optimization problem in the equation (\ref{eq-25}) becomes
\begin{equation}\label{eq-26}
\begin{split}
<\bm{B}>&=\arg\min_{\bm{B}}\textrm{Tr}((\bm{I}-\bm{B}\bm{S})^{T}\bm{K}(\bm{I}-\bm{B}\bm{S}))\\
     &\textrm{s.t.}\ ||\bm{\phi}(\bm{X})\bm{B}_{\cdot j}||_{2}^{2}\leq1,\ j=1,2,\cdots,k
\end{split}
\end{equation}

The problem (\ref{eq-26}) can be easily solved wtih the Lagrangian multiplier method:
\begin{equation}
\begin{split}
<\bm{B}>&=\arg\min_{\bm{B}}\textrm{Tr}((\bm{I}-\bm{B}\bm{S})^{T}\bm{K}(\bm{I}-\bm{B}\bm{S}))\\
   &+\sum_{i=1}^{k}\lambda_{i}(||\bm{\phi}(\bm{X})\bm{B}_{\cdot i}||_{2}^{2}-1)\nonumber
\end{split}
\end{equation}
where $\lambda_{i}\geq0,\ i=1,2,\cdots,k$ are Lagrangian parameters. Then we have:
\begin{equation}\label{eq-28}
\bm{B}=\bm{S}^{T}(\bm{S}\bm{S}^{T}+\bm{\Lambda})^{-1}
\end{equation}
where $\bm{\Lambda}$ is a diagonal matrix whose diagonal elements are $\lambda_{i},\ i=1,2,\cdots,k$.

\subsection{Optimization of the sparse coefficient $\bm{S}$}
Removing irrelevant terms in the equation (\ref{eq-25}), the optimization problem for $\bm{S}$ becomes:
\begin{equation}\label{eq-29}
\begin{split}
<\bm{S}>&=\arg\min_{\bm{S}}\textrm{Tr}((\bm{I}-\bm{B}\bm{S})^{T}\bm{K}(\bm{I}-\bm{B}\bm{S}))+\lambda\parallel \bm{S}\parallel_{1}\\
   &+\alpha\parallel \bm{H}-\bm{W}\bm{S}\parallel_{2}^{2}
\end{split}
\end{equation}

The equation (\ref{eq-29}) can be solved with the standard ADMM method by introducing an auxiliary variable $\bm{Z}$ and let $\bm{S}=\bm{Z}$. The optimization problem (\ref{eq-29}) can be written as:
\begin{equation}\label{eq-29-1}
\begin{split}
<\bm{Z},\bm{S}>&=\arg\min_{\bm{Z},\bm{S}}\textrm{Tr}((\bm{I}-\bm{B}\bm{S})^{T}\bm{K}(\bm{I}-\bm{B}\bm{S}))+\lambda\parallel \bm{Z}\parallel_{1}\\
 &+\alpha\parallel \bm{H}-\bm{W}\bm{S}\parallel_{2}^{2}\\
 &\textrm{s.t.}\ \bm{S}=\bm{Z},
\end{split}
\end{equation}
which can be reformulated as an unconstraint problem:
\begin{equation}\label{eq-29-2}
\begin{split}
<\bm{Z},\bm{S}>&=\arg\min_{\bm{Z},\bm{S}}\textrm{Tr}((\bm{I}-\bm{B}\bm{S})^{T}\bm{K}(\bm{I}-\bm{B}\bm{S}))+\lambda\parallel \bm{Z}\parallel_{1}\\
 &+\alpha\parallel \bm{H}-\bm{W}\bm{S}\parallel_{2}^{2}+\frac{\rho}{2}||\bm{S}-\bm{Z}||^{2}_{2}+\bm{y}^{T}(\bm{S}-\bm{Z}),
\end{split}
\end{equation}
where $\rho$ is a parameter to relax the constraint, while $\bm{y}$ is the Lagrangian multiplier. Then the problem in (\ref{eq-29-2}) can be decomposed into three subproblems:
\begin{eqnarray}\label{eq-29-3}
\left\{\begin{array}{ll}
     \bm{Z}^{k+1}=\arg\min_{\bm{Z}}\lambda||\bm{Z}||_{1}+\frac{\rho}{2}||\bm{S}^{k}-\bm{Z}||_{2}^{2}\\
     \ \ \ \ \ \ \ \ \ \ +(\bm{y}^{k})^{T}(\bm{S}^{k}-\bm{Z})\\
     \bm{S}^{k+1}=\arg\min_{\bm{S}}\textrm{Tr}((\bm{I}-\bm{B}\bm{S})^{T}\bm{K}(\bm{I}-\bm{B}\bm{S}))\\
     \ \ \ \ \ \ \ \ \ \ +\alpha\parallel \bm{H}-\bm{W}\bm{S}\parallel_{2}^{2}+\frac{\rho}{2}||\bm{S}-\bm{Z}^{k+1}||_{2}^{2}\\
     \ \ \ \ \ \ \ \ \ \ +(\bm{y}^{k})^{T}(\bm{S}-\bm{Z}^{k+1})\\
     \bm{y}^{k+1}=\bm{y}^{k}+\rho(\bm{S}^{k+1}-\bm{Z}^{k+1})
     \end{array}\right.
\end{eqnarray}

The first subproblem in (\ref{eq-29-3}) has a closed-form solution \cite{donoho1995noising}:
\begin{equation}\label{eq-29-4}
\bm{Z}^{k+1}=\bm{T}_{\lambda/\rho}(\bm{S}^{k}+\bm{y}^{k}/\rho)
\end{equation}
where the soft-thresholding operator $\bm{T}_{\lambda}(x)$ is defined as:
\begin{eqnarray}
\bm{T}_{\lambda}(x)=\left\{\begin{array}{ll}
     sgn(x)(|x|-\lambda),&|x|>\lambda\\
     0,&|x|\leq\lambda
     \end{array}\right.\nonumber
\end{eqnarray}
The second subproblem is a quadratic problem with solution:
\begin{equation}\label{eq-29-5}
\begin{split}
\bm{S}^{k+1}&=(2\bm{B}^{T}\bm{K}\bm{B}+2\alpha \bm{W}^{T}\bm{W}+\rho \bm{I})^{-1}(2\bm{B}^{T}\bm{K}+2\alpha \bm{W}^{T}\bm{H}\\
       &+\rho \bm{Z}^{k+1}-\bm{y}^{k})
\end{split}
\end{equation}

The algorithm to solve the problem (\ref{eq-29}) is summarized in the Algorithm \ref{alg2}.
\begin{algorithm}[!t]
\caption{Kernel sparse representation algorithm based on ADMM (KSR-ADMM)}
\label{alg2}
\algsetup{linenodelimiter=.}
\begin{algorithmic}[1]
\REQUIRE The kernel matrix $K$, dictionary associated matrix $B$, classifier matrix $W$ and predicted label matrix $H$, parameters $\lambda$, $\alpha$ and $\rho$.
\STATE Initialize $Z_{0}$, $S_{0}$ and $y_{0}$.
\FOR{$i=1$ to $J_{2}$}
\STATE Update $Z$ according to the equation (\ref{eq-29-4});
\STATE Update $S$ according to (\ref{eq-29-5});
\STATE Update $y$ according to the last equation in (\ref{eq-29-3}).
\ENDFOR
\ENSURE The sparse coefficient $S$.
\end{algorithmic}
\end{algorithm}

\subsection{Optimization of the classifier matrix $\bm{W}$}
The optimization problem for $\bm{W}$ is:
\begin{equation}\label{eq-30}
\begin{split}
<\bm{W}>&=\arg\min_{\bm{W}}\alpha||\bm{H}-\bm{W}\bm{S}||_{2}^{2},\\
 &\textrm{s.t.}\ ||\bm{W}_{\cdot m}||_{2}^{2}\leq1,\ m=1,2,\cdots,k
\end{split}
\end{equation}
and can be solved by the Lagrange Dual algorithm \cite{lee2006efficient}.

\subsection{Optimization of the predicted label matrix $\bm{H}$}
Finally, we can obtain the predicted label matrix $\bm{H}$ by solving the following problem:
\begin{equation}
\begin{split}
\bm{H}&=\arg\min_{\bm{H}}\alpha\parallel \bm{H}-\bm{W}\bm{S}\parallel_{2}^{2}+\textrm{Tr}(\bm{H}\tilde{\bm{L}}\bm{H}^{T})\\
 &+\gamma \textrm{Tr}((\bm{H}-\bm{F})\bm{U}(\bm{H}-\bm{F})^{T})\nonumber
\end{split}
\end{equation}

The analytical solution to the above problem is:
\begin{equation}\label{eq-31}
\bm{H}=(2\alpha \bm{W}\bm{S}+2\gamma \bm{F}\bm{U})(2\alpha \bm{I}+\tilde{\bm{L}}+\tilde{\bm{L}}^{T}+2\gamma \bm{U})^{-1}
\end{equation}

To summarize, the whole algorithm for the kernel method is described in the Algorithm \ref{alg3}.
\begin{algorithm}[!t]
\caption{Kernel semi-supervised sparse representation with graph regularization algorithm (KSSRGR)}
\label{alg3}
\algsetup{linenodelimiter=.}
\begin{algorithmic}[1]
\REQUIRE The image data $\bm{X}$, training label matrix $\bm{F}$ and label regularization matrix $\bm{U}$.
\STATE Initialize $\bm{B}_{0}$, $\bm{S}_{0}$, $\bm{W}_{0}$ and $\bm{H}_{0}$ similarly as the Algorithm \ref{alg1}.
\FOR{$i=1$ to $J_{3}$}
\STATE Update the dictionary associated matrix $\bm{B}$ according to (\ref{eq-28}) ;
\STATE Update the sparse coefficient $\bm{S}$ according to the Algorithm \ref{alg2};
\STATE Update the classifier matrix $\bm{W}$ using the Lagrange Dual algorithm \cite{lee2006efficient};
\STATE Update the label matrix $\bm{H}$ according to (\ref{eq-31}).
\ENDFOR
\ENSURE The dictionary associated matrix $\bm{B}$, sparse coefficient $\bm{S}$, classifier matrix $\bm{W}$ and predicted label matrix $\bm{H}$.
\end{algorithmic}
\end{algorithm}

\section{Experimental results and analysis}
In this section, we evaluate our approach on several standard databases, including the Extended YaleB database \cite{georghiades2001few}, the AR face database \cite{martinez1998ar} and the fifteen scene database \cite{lazebnik2006beyond}. We compare our algorithm with some popular algorithms, including unsupervised algorithms: the K-SVD algorithm \cite{aharon2006img}, the discriminative K-SVD (D-KSVD) algorithm \cite{zhang2010discriminative}, the sparse representation-based classification algorithm (SRC) \cite{wright2009robust}, and the locality-constrained linear coding algorithm (LLC) \cite{wang2010locality}, the supervised algorithm: LC-KSVD algorithm \cite{jiang2013label}, and the semi-supervised sparse coding algorithm (SSSC) \cite{wang2014semi}. The experiments are performed on a laptop with an i7-4720HQ 2.6-GHz CPU and 16-GB RAM running MATLAB 2018a.

\subsection{The Extended YaleB database}
\begin{figure}[!t]
\centering
\includegraphics[width=8.6cm,height=3.3cm]{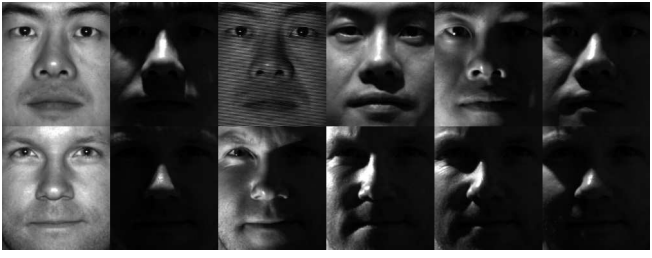}
\caption{Examples of the Extended YaleB face database}\label{Fig1}
\end{figure}

The Extended YaleB database consists of $2,414$ frontal face images in total, with $64$ images per class for $38$ people \cite{georghiades2001few}. The size of the original images in the database is $192\times168$. The examples of the database are shown in Fig.~\ref{Fig1}. As can be seen that the images vary with different illumination conditions and expressions. Therefore, the database is challenging for classification. In the experiments, half of the images from each category are randomly selected for training and the rest for testing. We project the original images to $504$-dimensional vectors with a random matrix. The dictionary size is set to $570$, corresponding to $15$ images for each class on average. Unlike algorithms, such as LC-KSVD and SRC, which have explicit correspondences between the dictionary columns and the labels of people, the proposed SSRGR algorithm does not enforce such constraints. Instead, we are trying to make full use of the underlying information contained in the unlabeled data to achieve better performance. The parameters in the proposed SSRGR and KSSRGR algorithms are tuned with $5$-fold cross-validation in a way varying one while keeping the others fixed. In the experiment, we set $\alpha=0.2$, $\gamma=0.06$ and $\lambda=1.2$ for the SSRGR algorithm and $\alpha=0.07$, $\gamma=0.0003$ and $\lambda=0.003$ for the KSSRGR algorithm. The sparsity in the algorithms K-SVD, D-KSVD, LLC, and LC-KSVD is set to $30$ as in \cite{jiang2013label} and the SSSC algorithm keeps the same sparsity parameter as the SSRGR algorithm.

\begin{table}[!t]
\caption{Classification results on the Extended YaleB database}
\label{table1}
\centering
\begin{tabular}{c|c}
\toprule
\multicolumn{1}{c|}{\bf Methods} & \multicolumn{1}{c}{\bf Accuracy(\%)}   \\
\midrule
K-SVD \cite{aharon2006img}                    & 91.65                     \\
\hline
D-KSVD \cite{zhang2010discriminative}         & 92.57                     \\
\hline
SRC \cite{wright2009robust}                   & 90.15                     \\
\hline
LLC \cite{wang2010locality}                   & 82.20                     \\
\hline
LC-KSVD \cite{jiang2013label}                 & 95.00                     \\
\hline
SSSC \cite{wang2014semi}                      & 95.41                     \\
\hline
SSRGR ($\beta_{1}=0$, $\beta_{2}=0$, $\beta_{3}=0$)    & 93.41            \\
\hline
SSRGR ($\beta_{1}\neq0$, $\beta_{2}=0$, $\beta_{3}=0$)    & 95.83         \\
\hline
SSRGR ($\beta_{1}\neq0$, $\beta_{2}\neq0$, $\beta_{3}\neq0$)   & 95.99    \\
\hline
KSSRGR ($\beta_{1}=0$, $\beta_{2}=0$, $\beta_{3}=0$)    & 95.24           \\
\hline
KSSRGR ($\beta_{1}\neq0$, $\beta_{2}=0$, $\beta_{3}=0$)     & 96.49       \\
\hline
KSSRGR ($\beta_{1}\neq0$, $\beta_{2}\neq0$, $\beta_{3}\neq0$)  & {\bf 96.66}   \\
\bottomrule
\end{tabular}
\end{table}


\begin{figure}[!t]
\centering
\includegraphics[width=8.8cm,height=4.8cm]{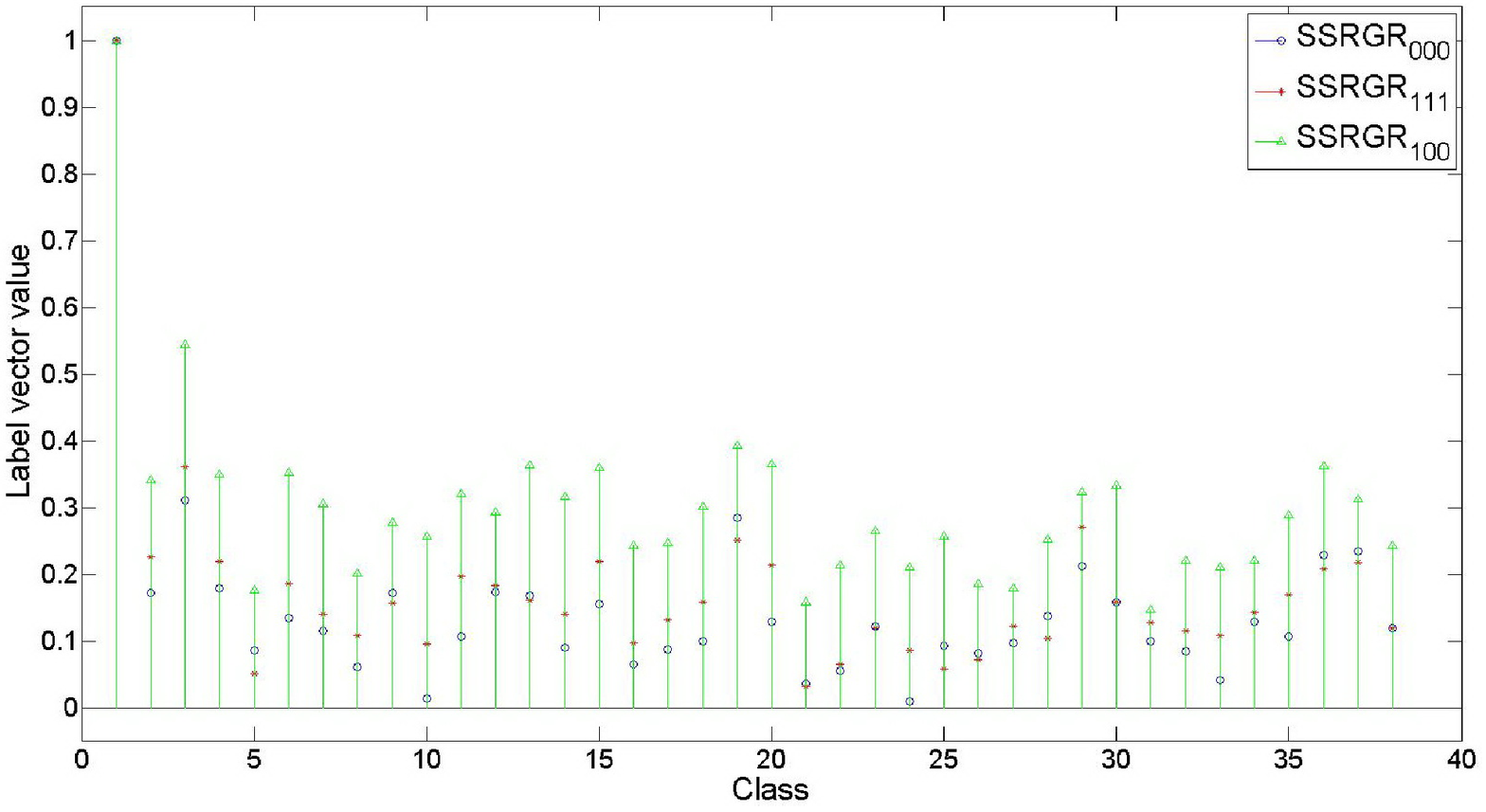}
\caption{The corresponding label vector values for an image from the class $1$} \label{Fig1-1}
\end{figure}

\begin{figure}[!t]
\centering
\includegraphics[width=8.8cm,height=4.8cm]{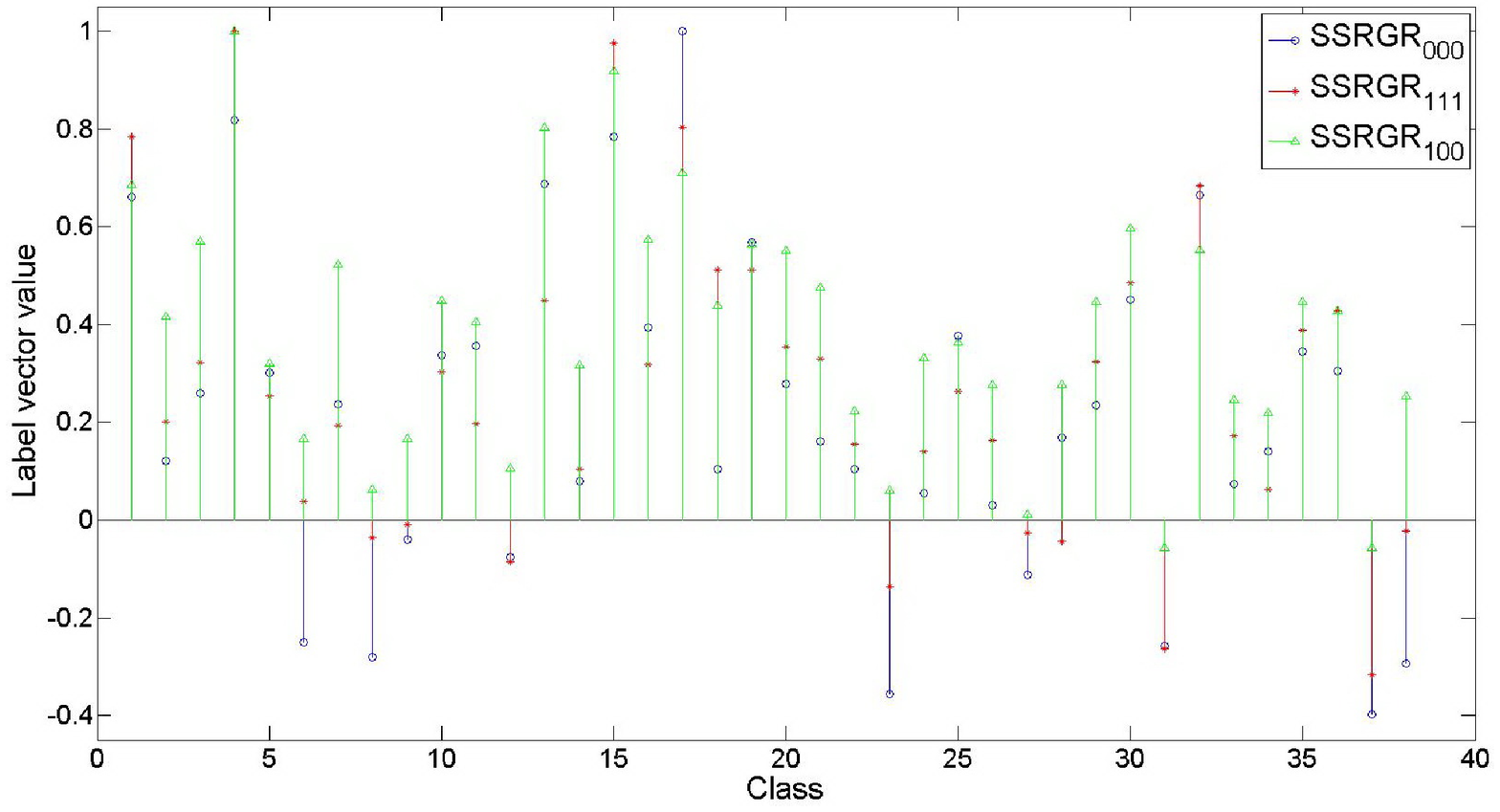}
\caption{The corresponding label vector values for an image from the class $4$} \label{Fig1-2}
\end{figure}

\begin{figure}[!t]
\centering
\includegraphics[width=8.8cm,height=4.8cm]{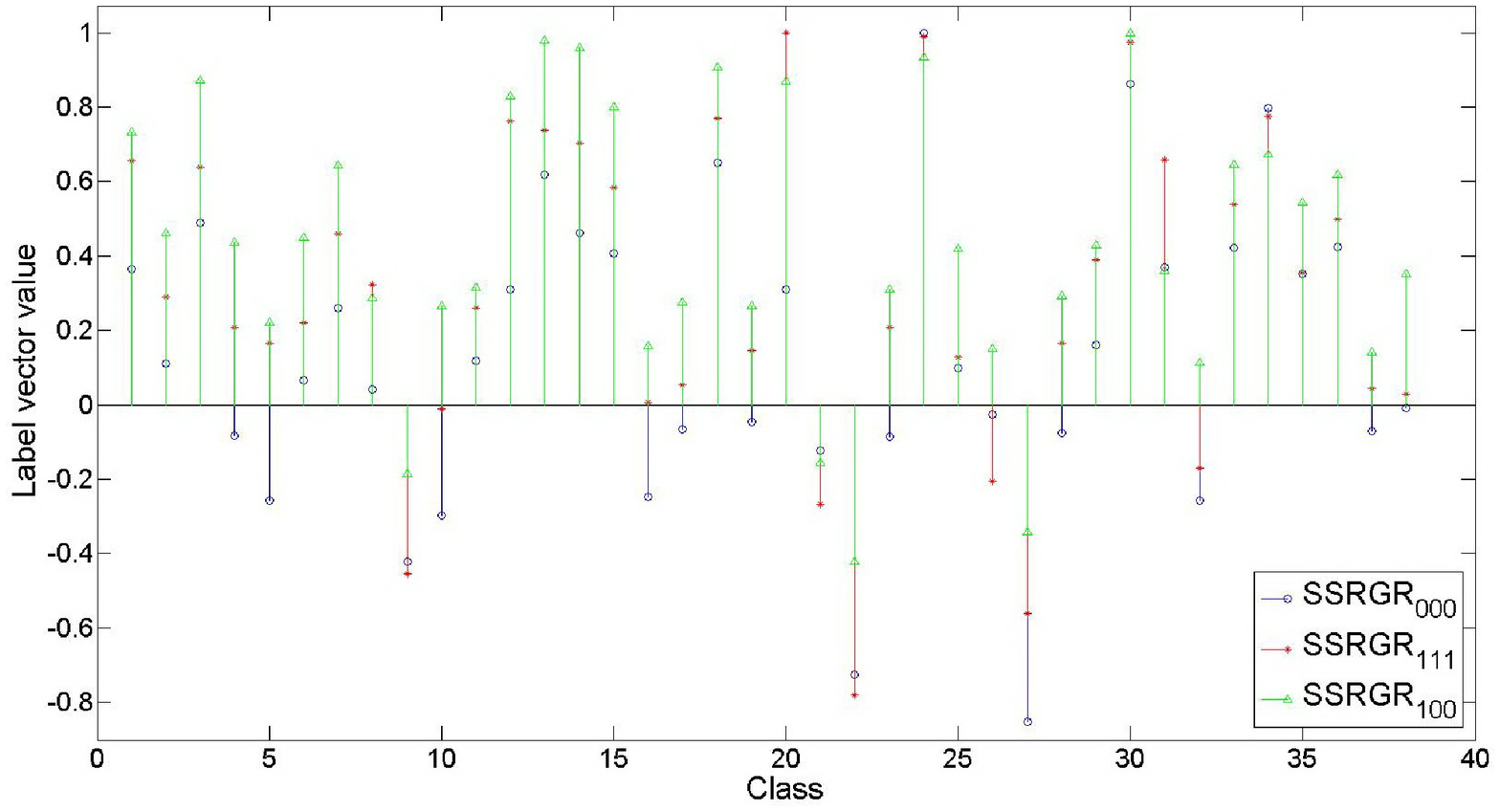}
\caption{The corresponding label vector values for an image from the class $20$} \label{Fig1-3}
\end{figure}

\begin{figure}[!t]
\centering
\includegraphics[width=8.8cm,height=4.8cm]{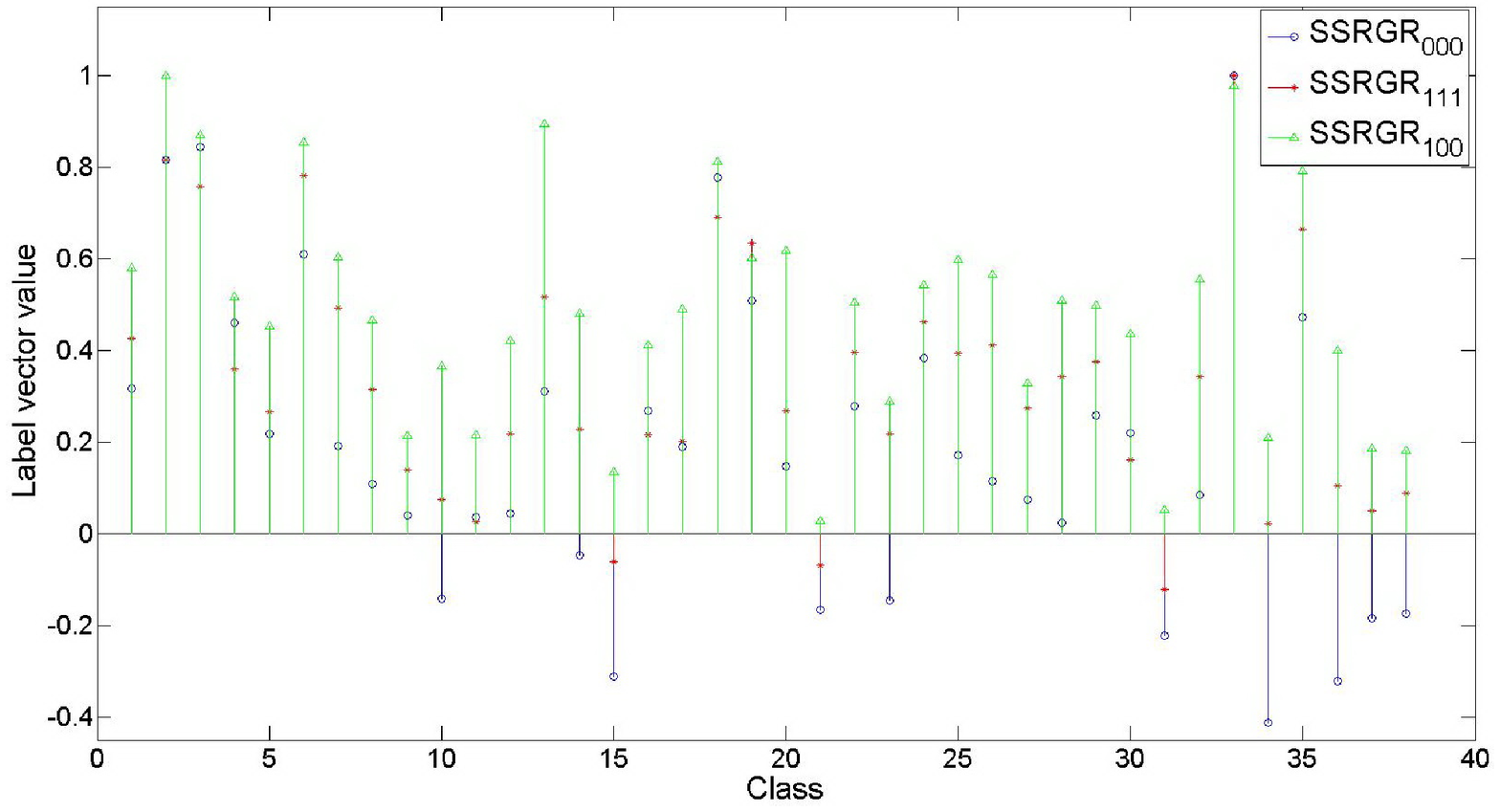}
\caption{The corresponding label vector values for an image from the class $33$} \label{Fig1-4}
\end{figure}

\begin{figure}[!t]
\centering
\includegraphics[width=8.8cm,height=4.8cm]{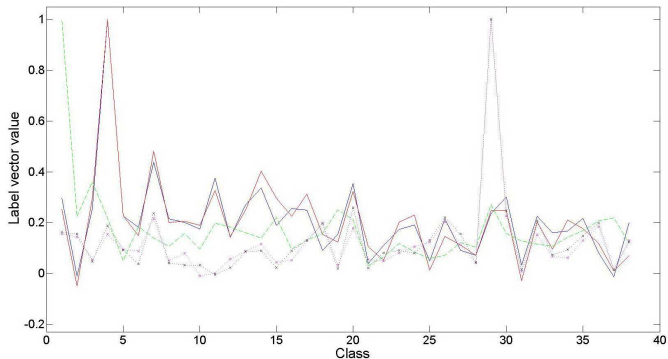}
\caption{The corresponding label vectors of images from different classes. The blue and red solid lines correspond to the images from the class $4$, the purple and black dash lines with ``$\times$" dots correspond to the images from the class $29$, while the green line corresponds to the images from the class $1$.} \label{Fig1-5}
\end{figure}

The experimental results are demonstrated in Table \ref{table1}. The parameters $\beta_{1}$, $\beta_{2}$ and $\beta_{3}$ are the weights for the three graph regularizations, i.e., the global manifold graph, the within-class graph and the between-class graph regularizations respectively. To evaluate the effectiveness of the proposed within-class and between-class graphs for the semi-supervised learning, we run the proposed SSRGR and KSSRGR algorithms with $\beta_{1}$, $\beta_{2}$ and $\beta_{3}$ being equal to $0$ or not respectively. As can be seen that the proposed SSRGR and KSSRGR algorithms always outperform the other algorithms when all the parameters $\beta_{1}$, $\beta_{2}$ and $\beta_{3}$ are nonzero. However, the performance of the SSRGR ang KSSRGR algorithms will decrease when $\beta_{1}\neq0$ and $\beta_{2}=\beta_{3}=0$, indicating that the proposed within-class and between-class graph regularizations are effective for promoting the discriminative ability for the proposed algorithms. Furthermore, the proposed algorithms perform poorer when all the three parameters are zero, which verifies the significance of the global manifold graph regularization for keeping the structure information of the whole data. Though no correspondences between the dictionary columns and the labels of people are enforced, the proposed SSRGR algorithm still achieves better performance than the LC-KSVD and SRC algorithms. Therefore, we can concluded that the underlying information included in the unlabeled data are beneficial for classification.

To further demonstrate the significant role the three graph regularizations play in the proposed algorithms, we present the element-wise values of the label vectors in label matrix $\bm{H}$ in Fig.~\ref{Fig1-1}, Fig.~\ref{Fig1-2}, Fig.~\ref{Fig1-3}, Fig.~\ref{Fig1-4}, and Fig.~\ref{Fig1-5}. In these figures, the blue circle points represent the vectors obtained by the SSRGR algorithm performed with $\beta_{1}=\beta_{2}=\beta_{3}=0$, the green triangle points represent the vectors obtained with $\beta_{1}\neq0$ and $\beta_{2}=\beta_{3}=0$, while the red star points represent the vectors obtained with all the three parameters being nonzero. For the convenience, we denote the three cases with $SSRGR_{000}$, $SSRGR_{100}$ and $SSRGR_{111}$ respectively. For fair comparison, we normalize each label vector with its maximal value. In Fig.~\ref{Fig1-1}, all the three label vectors peak at the first position. Since the position where the maximal value achieves indicates the class number of the corresponding image, the image is correctly classified with in all three cases. It can be observed that most element-wise values of $SSRGR_{000}$ are smaller than those of $SSRGR_{100}$ and $SSRGR_{111}$, indicating that the graph regularizations will regularize the values of the label vectors so as to obtain the correct classification resuts. It should be noted that most values of the label vectors of $SSRGR_{111}$ are smaller than those of $ASSRGR_{100}$ due to the influence of the within-class and between-class graph regularizations. The image from the class $4$ is incorrectly classified to the class $17$ with $SSRGR_{000}$ in Fig. \ref{Fig1-2}. By contrast, both $SSRGR_{100}$ and $SSRGR_{111}$ provide correct classification results. In Fig. \ref{Fig1-3}, only $SSRGR_{111}$ classifies the images to the class $20$ correctly. Though the maximal values of the label vector with $SSRGR_{000}$ are suppressed to a relatively smaller values with $SSRGR_{100}$, the value at the position of $30$ becomes maximum. By contrast, the $SSRGR_{111}$ not only suppresses the maximal values of the $SSRGR_{000}$, but also keeps the values at the position of $20$ being the maximum. In Fig. \ref{Fig1-4}, the $SSRGR_{100}$ achieves maximum at the second position of the label vector wrongly, while the $SSRGR_{111}$ increases the values of the label vector of $SSRGR_{000}$ and makes the value at the position of $33$ be the largest. Fig. \ref{Fig1-5} shows the label vectors obtained with $SSRGR_{111}$ for images from different classes. We can observe that the patterns of the label vectors for images from the same class are similar, while vary greatly for images from different classes. Therefore, we can conclude from the above analysis that the three graph regularizations are able to extract beneficial structure information from the whole data for classification.

\begin{table}[!t]
\caption{The computing time of the algorithms to classify $1,198$ testing images in the Extended YaleB database}
\label{table1-1}
\centering
\begin{tabular}{c|c}
\toprule
\multicolumn{1}{c|}{\bf Methods} & \multicolumn{1}{c}{\bf Time(s)}   \\
\midrule
K-SVD                            & 385.54                           \\
\hline
D-KSVD                           & 798.97                           \\
\hline
SRC                              & 17.05                            \\
\hline
LC-KSVD                          & 38.16                            \\
\hline
SSSC                             & 1293.61                          \\
\hline
SSRGR                            & 35.66                            \\
\hline
KSSRGR                           & 89.44                            \\
\bottomrule
\end{tabular}
\end{table}

In addition, we compare the algorithms regarding the computational time to classify the $1,198$ testing images and the result is shown in the Table \ref{table1-1}. The D-KSVD algorithm achieves higher accuracy than the K-SVD algorithm but costs more time, as it needs extra time to learn a data-driven classifier. The LC-KSVD algorithm is of much more computational efficiency than the aforementioned two algorithms while obtaining better classification performance. As the SRC algorithm utilizes the training data as the dictionary for the sparse coding process, it needs the least time to classify the testing images. Although the SSSC algorithm can obtain comparable accuracy with the proposed SSRGR algorithm, however, it takes the most time to complete the classification task. By contrast, the proposed SSRGR algorithm consumes the least time compared with the algorithms except the SRC algorithm. The KSSRGR algorithm achieves the best accuracy while only needs more time than the SSRGR and LC-KSVD algorithms. In summary, the proposed methods can obtain the best classification result with highly computational efficiency.

\subsection{The AR face database}
\begin{figure}[!t]
\centering
\includegraphics[width=8.7cm,height=2.5cm]{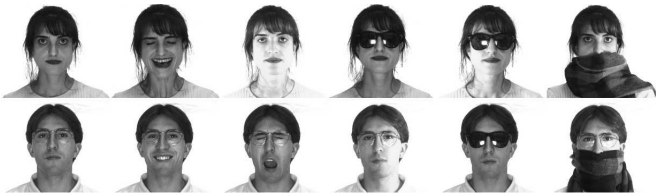}
\caption{Examples of the AR face database} \label{Fig2}
\end{figure}

The AR face database \cite{martinez1998ar} is a collection of $126$ individuals with $26$ color images taken during two sessions for each subject. The total number of the images is over $4,000$. Compared with the Extended YaleB face database, the images in the AR database contain more facial variations, which vary greatly with respect to the facial expressions, illumination conditions and occlusions induced by the sunglasses and scarves, as illustrated in Fig.~\ref{Fig2}. Following the settings in \cite{jiang2013label}, we utilize a subset of the database including $2,600$ images from $50$ male individuals and $50$ female individuals. The original images with the size of $165\times120$ are projected to $540$-dimensional vectors with a random matrix. We randomly select $5$, $10$, $15$ and $20$ images from each class respectively for training and the rest for testing. In the four different experiments, the total number of columns in the learned dictionary is always set to $500$, corresponding to $5$ images for each class.

\begin{figure}[!t]
\centering
\includegraphics[width=8.0cm,height=4.4cm]{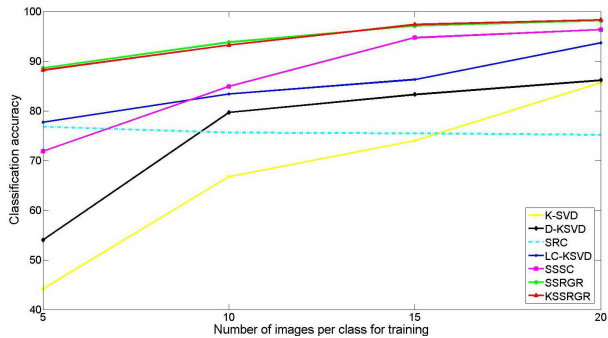}
\caption{Classification results with different number of training images on the AR face database} \label{Fig2-0}
\end{figure}

The classification results are illustrated in Fig.~\ref{Fig2-0}. As is shown that the proposed SSRGR and KSSRGR algorithms consistently achieve the best results. Particularly, when the training number is small, the proposed method tends to present greater advantage over the others. With the increasing of the number of the training images, the classification results of the proposed SSRGR and KSSRGR algorithms do not vary dramatically as the other algorithms. As the underlying structure information contained in the unlabeled data together with the supervised information of the labeled data are fully utilized, the proposed methods are able to extract discriminative features from the data and achieve excellent classification performances, even when the number of training images is small. Therefore, we can conclude that the utilization of the underlying information in the unlabeled data can enhance the classification performances of the supervised learning algorithms.

\begin{table}[!t]
\caption{The computing times for different algorithms to classify $600$ testing images in the AR face database}
\label{table2-1}
\centering
\begin{tabular}{c|c}
\toprule
\multicolumn{1}{c|}{\bf Methods} & \multicolumn{1}{c}{\bf Time(s)}   \\
\midrule
K-SVD                            & 610.15                            \\
\hline
D-KSVD                           & 1286.68                           \\
\hline
SRC                              & 9.46                              \\
\hline
LC-KSVD                          & 47.35                             \\
\hline
SSSC                             & 6542.11                           \\
\hline
SSRGR                            & 35.81                             \\
\hline
KSSRGR                           & 87.43                             \\
\bottomrule
\end{tabular}
\end{table}

\begin{figure}[!t]
\centering
\includegraphics[width=7.0cm,height=4.0cm]{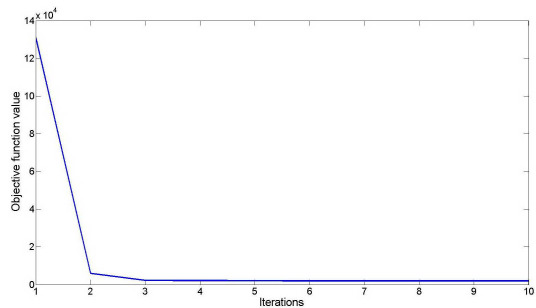}
\caption{The convergence of SSRGR algorithm}\label{Fig2-1}
\end{figure}

\begin{figure}[!t]
\centering
\includegraphics[width=7.0cm,height=3.9cm]{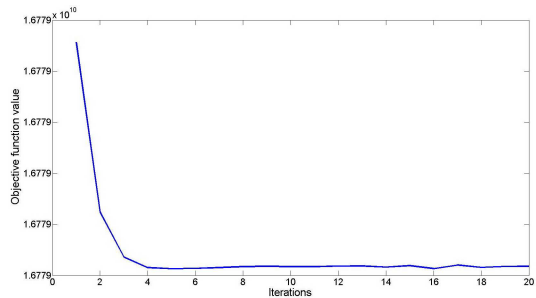}
\caption{The convergence of KSSRGR algorithm}\label{Fig2-2}
\end{figure}

We compare the computational time of algorithms when the number of testing images is $600$ in Table \ref{table2-1}. As can be seen that the proposed SSRGR is the fastest among the algorithms except the SRC algorithm, while the KSSRGR algorithm which achieves the best classification accuracy is only slower than the SSRGR and LC-KSVD algorithms. Fig. \ref{Fig2-1} and Fig. \ref{Fig2-2} demonstrate the convergency of the SSRGR and KSSRGR algorithms respectively. As is shown that both algorithms can converge in a few iterations. It should be noted that although there are two iteration loops in Algorithm \ref{alg1} and \ref{alg3}, the number of the inner iteration is always set to $1$ in the experiments, making the algorithms highly efficient.

\subsection{The fifteen scene database}
\begin{figure}[!t]
\centering
\includegraphics[width=8.7cm,height=2.5cm]{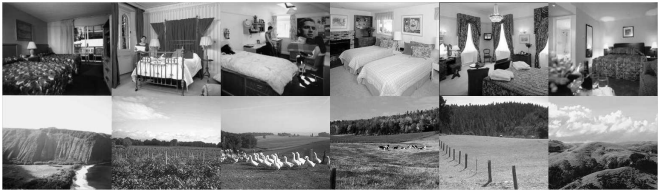}
\caption{Examples of bedroom and open country in the fifteen scene database} \label{Fig3}
\end{figure}

The fifteen scene database consists of fifteen different natural indoor and outdoor scenes, including office, kitchen, living room, bedroom, store, industrial, tall building, inside city, street, highway, coast, open country, mountain, forest, and suburb, which is first introduced in \cite{lazebnik2006beyond}. The images in the database are of average size $250\times300$, with $210$ to $400$ images per class. The examples of the bedroom and open country scenes in the database are shown in Fig. \ref{Fig3}. Following the common experimental settings, $100$ images from each class are randomly selected for training and the rest for testing. The dictionary is set to have $450$ atoms. The spatial pyramid features of the images are employed for the classification task.

\begin{table}[!t]
\caption{Classification results on the fifteen scene database using the spatial pyramid features}
\label{table3}
\centering
\begin{tabular}{c|c}
\toprule
\multicolumn{1}{c|}{\bf Methods} & \multicolumn{1}{c}{\bf Accuracy(\%)}   \\
\midrule
K-SVD \cite{aharon2006img}            & 86.70                                  \\
\hline
D-KSVD \cite{zhang2010discriminative}            & 89.10                                  \\
\hline
SRC \cite{wright2009robust}              & 91.80                                  \\
\hline
LLC \cite{wang2010locality}                & 89.20                                  \\
\hline
LC-KSVD  \cite{jiang2013label}          & 92.90                                  \\
\hline
Lazebnik\cite{lazebnik2006beyond}           & 81.40                                  \\
\hline
Gemert\cite{van2008kernel}            & 76.70                                   \\
\hline
Yang\cite{yang2009linear}                & 80.30                                  \\
\hline
gao\cite{gao2010local}                  &89.70                                   \\
\hline
Lian\cite{lian2010max}                & 86.40                                   \\
\hline
Boureau\cite{boureau2010learning}             & 84.30                                   \\
\hline
SSSC \cite{wang2014semi}               & 97.52                                  \\
\hline
SSRGR                            & 96.45                                  \\
\hline
KSSRGR                           & {\bf 97.89}                            \\
\bottomrule
\end{tabular}
\end{table}

We compare the proposed approach with popular algorithms, including K-SVD \cite{aharon2006img}, D-KSVD \cite{zhang2010discriminative}, SRC \cite{wright2009robust}, LLC \cite{wang2010locality}, LC-KSVD \cite{jiang2013label}, SSSC \cite{wang2014semi} and other methods \cite{yang2009linear,lazebnik2006beyond,van2008kernel,gao2010local,lian2010max,boureau2010learning}. In the experiments, we set the LLC method to have $30$ local bases. The classification results are demonstrated in Table~\ref{table3}. As can be seen that the SSSC algorithm and the proposed SSRGR and KSSRGR algorithms achieve much better results than the others. Compared with two face databases, the images in the fifteen scene database vary greatly between different classes and within the same classes. Consequently, more information is needed to fulfill the challenging classification task. Semi-supervised methods is able to exploit the structure information in the unlabeled data. As a result, they tend to achieve better performances. Although the performance of the SSSC algorithm is slightly better than that of the proposed SSRGR algorithm, it is lower than that of the KSSRGR algorithm. It should also be noted that the SSSC algorithm is much less efficient than both the SSRGR and KSSRGR algorithms.

\section{Conclusion and future work}
In this paper, we propose a novel discriminative semi-supervised sparse representation with graph constraint method for image classification. The proposed method aims to utilize the information included in the unlabeled data to achieve better performances. We combine the classifier learning process with the sparse coding process to learn a data-driven linear classifier. To extract the underlying information both in the labeled and unlabeled data, three graphs, the global manifold structure graph, the within-class graph and the between-class graph, are constructed. Then, the constructed three graphs are employed to regularize the predicted label vectors to keep the same manifold structure as the original data and be discriminative from each other. Furthermore, we extend the proposed algorithm to its kernel version so as to classify data with nonlinear structure property. Consequently, two efficient algorithms are developed to solve the corresponding optimization problems. We evaluate the proposed algorithms on challenging databases  and demonstrated that the proposed algorithms can achieve superior performance compared with other popular algorithms. In the future, we may consider to apply the proposed methods to data with noise. In addition, seeking better methods to regularize the predicted label vectors to further improve the classification performance can be another possible research direction.

\ifCLASSOPTIONcaptionsoff
  \newpage
\fi



%


%

\bibliographystyle{./IEEEtran}
\bibliography{mybibfile}

\begin{thebibliography}{10}
\providecommand{\url}[1]{#1}
\csname url@samestyle\endcsname
\providecommand{\newblock}{\relax}
\providecommand{\bibinfo}[2]{#2}
\providecommand{\BIBentrySTDinterwordspacing}{\spaceskip=0pt\relax}
\providecommand{\BIBentryALTinterwordstretchfactor}{4}
\providecommand{\BIBentryALTinterwordspacing}{\spaceskip=\fontdimen2\font plus
\BIBentryALTinterwordstretchfactor\fontdimen3\font minus
  \fontdimen4\font\relax}
\providecommand{\BIBforeignlanguage}[2]{{%
\expandafter\ifx\csname l@#1\endcsname\relax
\typeout{** WARNING: IEEEtran.bst: No hyphenation pattern has been}%
\typeout{** loaded for the language `#1'. Using the pattern for}%
\typeout{** the default language instead.}%
\else
\language=\csname l@#1\endcsname
\fi
#2}}
\providecommand{\BIBdecl}{\relax}
\BIBdecl

\bibitem{hoyer2002non}
P.~O. Hoyer, ``Non-negative sparse coding,'' in \emph{Neural Networks for
  Signal Processing, 2002. Proceedings of the 2002 12th IEEE Workshop
  on}.\hskip 1em plus 0.5em minus 0.4em\relax IEEE, 2002, pp. 557--565.

\bibitem{lee2006efficient}
H.~Lee, A.~Battle, R.~Raina, and A.~Y. Ng, ``Efficient sparse coding
  algorithms,'' in \emph{Advances in neural information processing systems},
  2006, pp. 801--808.

\bibitem{mairal2009online}
J.~Mairal, F.~Bach, J.~Ponce, and G.~Sapiro, ``Online dictionary learning for
  sparse coding,'' in \emph{Proceedings of the 26th Annual International
  Conference on Machine Learning}.\hskip 1em plus 0.5em minus 0.4em\relax ACM,
  2009, pp. 689--696.

\bibitem{zheng2011graph}
M.~Zheng, J.~Bu, C.~Chen, C.~Wang, L.~Zhang, G.~Qiu, and D.~Cai, ``Graph
  regularized sparse coding for image representation,'' \emph{Image Processing,
  IEEE Transactions on}, vol.~20, no.~5, pp. 1327--1336, 2011.

\bibitem{yang2009linear}
J.~Yang, K.~Yu, Y.~Gong, and T.~Huang, ``Linear spatial pyramid matching using
  sparse coding for image classification,'' in \emph{Computer Vision and
  Pattern Recognition, 2009. CVPR 2009. IEEE Conference on}.\hskip 1em plus
  0.5em minus 0.4em\relax IEEE, 2009, pp. 1794--1801.

\bibitem{li2014sparse}
H.~Li, H.~Li, Y.~Wei, Y.~Tang, and Q.~Wang, ``Sparse-based neural response for
  image classification,'' \emph{Neurocomputing}, vol. 144, pp. 198--207, 2014.

\bibitem{li2018neural}
H.~Li, H.~Zhao, and H.~Li, ``Neural-response-based extreme learning machine for
  image classification,'' \emph{IEEE transactions on neural networks and
  learning systems}, vol.~30, no.~2, pp. 539--552, 2018.

\bibitem{li2013hierarchical}
H.~Li, Y.~Wei, L.~Li, and C.~P. Chen, ``Hierarchical feature extraction with
  local neural response for image recognition,'' \emph{Cybernetics, IEEE
  Transactions on}, vol.~43, no.~2, pp. 412--424, 2013.

\bibitem{wright2009robust}
J.~Wright, A.~Y. Yang, A.~Ganesh, S.~S. Sastry, and Y.~Ma, ``Robust face
  recognition via sparse representation,'' \emph{Pattern Analysis and Machine
  Intelligence, IEEE Transactions on}, vol.~31, no.~2, pp. 210--227, 2009.

\bibitem{aharon2006img}
M.~Aharon, M.~Elad, and A.~Bruckstein, ``K-svd: An algorithm for designing
  overcomplete dictionaries for sparse representation,'' \emph{Signal
  Processing, IEEE Transactions on}, vol.~54, no.~11, pp. 4311--4322, 2006.

\bibitem{jiang2013label}
Z.~Jiang, Z.~Lin, and L.~S. Davis, ``Label consistent k-svd: learning a
  discriminative dictionary for recognition,'' \emph{Pattern Analysis and
  Machine Intelligence, IEEE Transactions on}, vol.~35, no.~11, pp. 2651--2664,
  2013.

\bibitem{guha2012learning}
T.~Guha and R.~K. Ward, ``Learning sparse representations for human action
  recognition,'' \emph{Pattern Analysis and Machine Intelligence, IEEE
  Transactions on}, vol.~34, no.~8, pp. 1576--1588, 2012.

\bibitem{shrivastava2015multiple}
A.~Shrivastava, J.~K. Pillai, and V.~M. Patel, ``Multiple kernel-based
  dictionary learning for weakly supervised classification,'' \emph{Pattern
  Recognition}, vol.~48, no.~8, pp. 2667--2675, 2015.

\bibitem{mairal2009supervised}
J.~Mairal, J.~Ponce, G.~Sapiro, A.~Zisserman, and F.~R. Bach, ``Supervised
  dictionary learning,'' in \emph{Advances in neural information processing
  systems}, 2009, pp. 1033--1040.

\bibitem{zhang2010discriminative}
Q.~Zhang and B.~Li, ``Discriminative k-svd for dictionary learning in face
  recognition,'' in \emph{Computer Vision and Pattern Recognition (CVPR), 2010
  IEEE Conference on}.\hskip 1em plus 0.5em minus 0.4em\relax IEEE, 2010, pp.
  2691--2698.

\bibitem{chapelle2006semi}
O.~Chapelle, B.~Sch{\"o}lkopf, A.~Zien \emph{et~al.}, ``Semi-supervised
  learning,'' 2006.

\bibitem{cai2007semi}
D.~Cai, X.~He, and J.~Han, ``Semi-supervised discriminant analysis,'' in
  \emph{Computer Vision, 2007. ICCV 2007. IEEE 11th International Conference
  on}.\hskip 1em plus 0.5em minus 0.4em\relax IEEE, 2007, pp. 1--7.

\bibitem{yang2012semi}
S.~Yang, X.~Wang, L.~Yang, Y.~Han, and L.~Jiao, ``Semi-supervised action
  recognition in video via labeled kernel sparse coding and sparse l 1 graph,''
  \emph{Pattern Recognition Letters}, vol.~33, no.~14, pp. 1951--1956, 2012.

\bibitem{wang2014semi}
J.~J.-Y. Wang and X.~Gao, ``Semi-supervised sparse coding,'' in \emph{Neural
  Networks (IJCNN), 2014 International Joint Conference on}.\hskip 1em plus
  0.5em minus 0.4em\relax IEEE, 2014, pp. 1630--1637.

\bibitem{yu2006active}
K.~Yu, J.~Bi, and V.~Tresp, ``Active learning via transductive experimental
  design,'' in \emph{Proceedings of the 23rd international conference on
  Machine learning}.\hskip 1em plus 0.5em minus 0.4em\relax ACM, 2006, pp.
  1081--1088.

\bibitem{he2011nonnegative}
R.~He, W.~S. Zheng, B.~G. Hu, and X.~W. Kong, ``Nonnegative sparse coding for
  discriminative semi-supervised learning,'' in \emph{Computer Vision and
  Pattern Recognition (CVPR), 2011 IEEE Conference on}.\hskip 1em plus 0.5em
  minus 0.4em\relax IEEE, 2011, pp. 2849--2856.

\bibitem{zheng2013image}
H.~Zheng and H.~H. Ip, ``Image classification by iterative semi-supervised
  sparse coding,'' in \emph{Advances in Multimedia Information Processing--PCM
  2013}.\hskip 1em plus 0.5em minus 0.4em\relax Springer, 2013, pp. 485--496.

\bibitem{chapelle2009semi}
O.~Chapelle, B.~Scholkopf, and A.~Zien, ``Semi-supervised learning (chapelle,
  o. et al., eds.; 2006)[book reviews],'' \emph{IEEE Transactions on Neural
  Networks}, vol.~20, no.~3, pp. 542--542, 2009.

\bibitem{yu2012solving}
G.~Yu, G.~Sapiro, and S.~Mallat, ``Solving inverse problems with piecewise
  linear estimators: From gaussian mixture models to structured sparsity,''
  \emph{IEEE Transactions on Image Processing}, vol.~21, no.~5, pp. 2481--2499,
  2012.

\bibitem{joachims2006transductive}
T.~Joachims, ``Transductive support vector machines,'' \emph{Chapelle et
  al.(2006)}, pp. 105--118, 2006.

\bibitem{gong2015deformed}
C.~Gong, T.~Liu, D.~Tao, K.~Fu, E.~Tu, and J.~Yang, ``Deformed graph laplacian
  for semisupervised learning,'' \emph{IEEE transactions on neural networks and
  learning systems}, vol.~26, no.~10, pp. 2261--2274, 2015.

\bibitem{candes2006stable}
E.~J. Candes, J.~K. Romberg, and T.~Tao, ``Stable signal recovery from
  incomplete and inaccurate measurements,'' \emph{Communications on pure and
  applied mathematics}, vol.~59, no.~8, pp. 1207--1223, 2006.

\bibitem{donoho2006compressed}
D.~L. Donoho, ``Compressed sensing,'' \emph{Information Theory, IEEE
  Transactions on}, vol.~52, no.~4, pp. 1289--1306, 2006.

\bibitem{sugiyama2006local}
M.~Sugiyama, ``Local fisher discriminant analysis for supervised dimensionality
  reduction,'' in \emph{Proceedings of the 23rd international conference on
  Machine learning}.\hskip 1em plus 0.5em minus 0.4em\relax ACM, 2006, pp.
  905--912.

\bibitem{zhang2014hyperspectral}
L.~Zhang, L.~Zhang, D.~Tao, X.~Huang, and B.~Du, ``Hyperspectral remote sensing
  image subpixel target detection based on supervised metric learning,''
  \emph{Geoscience and Remote Sensing, IEEE Transactions on}, vol.~52, no.~8,
  pp. 4955--4965, 2014.

\bibitem{liu2010constrained}
W.~Liu, X.~Tian, D.~Tao, and J.~Liu, ``Constrained metric learning via distance
  gap maximization.'' in \emph{AAAI}, 2010.

\bibitem{zhou2004learning}
D.~Zhou, O.~Bousquet, T.~N. Lal, J.~Weston, and B.~Sch{\"o}lkopf, ``Learning
  with local and global consistency,'' \emph{Advances in neural information
  processing systems}, vol.~16, no.~16, pp. 321--328, 2004.

\bibitem{kandola2002learning}
J.~Kandola, N.~Cristianini, and J.~S. Shawe-taylor, ``Learning semantic
  similarity,'' in \emph{Advances in neural information processing systems},
  2002, pp. 657--664.

\bibitem{boyd2011distributed}
S.~Boyd, N.~Parikh, E.~Chu, B.~Peleato, and J.~Eckstein, ``Distributed
  optimization and statistical learning via the alternating direction method of
  multipliers,'' \emph{Foundations and Trends in Machine Learning}, vol.~3,
  no.~1, pp. 1--122, 2011.

\bibitem{golub1999tikhonov}
G.~H. Golub, P.~C. Hansen, and D.~P. O'Leary, ``Tikhonov regularization and
  total least squares,'' \emph{SIAM Journal on Matrix Analysis and
  Applications}, vol.~21, no.~1, pp. 185--194, 1999.

\bibitem{van2013design}
H.~Van~Nguyen, V.~M. Patel, N.~M. Nasrabadi, and R.~Chellappa, ``Design of
  non-linear kernel dictionaries for object recognition,'' \emph{Image
  Processing, IEEE Transactions on}, vol.~22, no.~12, pp. 5123--5135, 2013.

\bibitem{donoho1995noising}
D.~L. Donoho, ``De-noising by soft-thresholding,'' \emph{Information Theory,
  IEEE Transactions on}, vol.~41, no.~3, pp. 613--627, 1995.

\bibitem{georghiades2001few}
A.~S. Georghiades, P.~N. Belhumeur, and D.~Kriegman, ``From few to many:
  Illumination cone models for face recognition under variable lighting and
  pose,'' \emph{Pattern Analysis and Machine Intelligence, IEEE Transactions
  on}, vol.~23, no.~6, pp. 643--660, 2001.

\bibitem{martinez1998ar}
A.~M. Martinez, ``The ar face database,'' \emph{CVC Technical Report}, vol.~24,
  1998.

\bibitem{lazebnik2006beyond}
S.~Lazebnik, C.~Schmid, and J.~Ponce, ``Beyond bags of features: Spatial
  pyramid matching for recognizing natural scene categories,'' in
  \emph{Computer Vision and Pattern Recognition, 2006 IEEE Computer Society
  Conference on}, vol.~2.\hskip 1em plus 0.5em minus 0.4em\relax IEEE, 2006,
  pp. 2169--2178.

\bibitem{wang2010locality}
J.~Wang, J.~Yang, K.~Yu, F.~Lv, T.~Huang, and Y.~Gong, ``Locality-constrained
  linear coding for image classification,'' in \emph{Computer Vision and
  Pattern Recognition (CVPR), 2010 IEEE Conference on}.\hskip 1em plus 0.5em
  minus 0.4em\relax IEEE, 2010, pp. 3360--3367.

\bibitem{van2008kernel}
J.~C. van Gemert, J.~M. Geusebroek, C.~J. Veenman, and A.~W. Smeulders,
  ``Kernel codebooks for scene categorization,'' in \emph{Computer Vision--ECCV
  2008}.\hskip 1em plus 0.5em minus 0.4em\relax Springer, 2008, pp. 696--709.

\bibitem{gao2010local}
S.~Gao, I.~W. Tsang, L.~T. Chia, and P.~Zhao, ``Local features are not
  lonely--laplacian sparse coding for image classification,'' in \emph{Computer
  Vision and Pattern Recognition (CVPR), 2010 IEEE Conference on}.\hskip 1em
  plus 0.5em minus 0.4em\relax IEEE, 2010, pp. 3555--3561.

\bibitem{lian2010max}
X.~C. Lian, Z.~Li, B.~L. Lu, and L.~Zhang, ``Max-margin dictionary learning for
  multiclass image categorization,'' in \emph{Computer Vision--ECCV
  2010}.\hskip 1em plus 0.5em minus 0.4em\relax Springer, 2010, pp. 157--170.

\bibitem{boureau2010learning}
Y.~L. Boureau, F.~Bach, Y.~LeCun, and J.~Ponce, ``Learning mid-level features
  for recognition,'' in \emph{Computer Vision and Pattern Recognition (CVPR),
  2010 IEEE Conference on}.\hskip 1em plus 0.5em minus 0.4em\relax IEEE, 2010,
  pp. 2559--2566.

\end{thebibliography}

%

%
%
%




\end{document}